\documentclass[letterpaper]{article} 
\ifdefined\pdfminorversion
\fi
\ifdefined\pdfobjcompresslevel
\fi
\usepackage[preprint]{aaai2027}  
\usepackage[hyphens]{url}  
\usepackage{graphicx} 
\def\UrlFont{\rm}  
\usepackage{natbib}  
\usepackage{caption} 
\usepackage{listings}
\DeclareCaptionFormat{promptlisting}{%
  \rule{\linewidth}{0.5pt}\par
  \vspace{1pt}%
  #1#2#3\par
  \vspace{1pt}%
  \rule{\linewidth}{0.5pt}}
\newcommand{\promptlistingcaption}[2]{%
  \begingroup
  \captionof{lstlisting}{#1}\label{#2}%
  \endgroup}
\newcommand{\promptlistingend}{%
  \par\nobreak\noindent\rule{\columnwidth}{0.5pt}%
  \par\addvspace{4pt}}
\makeatletter
\AtBeginDocument{\let\lst@MakeCaption\@gobble}
\makeatother

\usepackage{algorithm}
\usepackage{algorithmic}

\usepackage{booktabs}
\usepackage{array}

\usepackage{amsmath}
\usepackage{amssymb}
\usepackage{amsthm}

\title{EviGraph: Evidence-Guided Autonomous Research Agents}

\author{
    Zhenjiang Ren\textsuperscript{\rm 1,\rm 2},
    Ruiji Li\textsuperscript{\rm 1},
    Xujing Zhang\textsuperscript{\rm 3},
    Ziliang Pang\textsuperscript{\rm 1},
    Shuo Ren\textsuperscript{\rm 1}\corresponding,
    Jiajun Zhang\textsuperscript{\rm 1,\rm 2,\rm 4}\corresponding
}

\affiliations{
    \textsuperscript{\rm 1}Institute of Automation, Chinese Academy of Sciences\\
    \textsuperscript{\rm 2}School of Artificial Intelligence, University of Chinese Academy of Sciences\\
    \textsuperscript{\rm 3}Hong Kong Baptist University  \textsuperscript{\rm 4}Wuhan AI Research\\
    \{renzhenjiang2024, liruiji2026, ziliang.pang, shuo.ren\}@ia.ac.cn, \\24267368@life.hkbu.edu.hk, jjzhang@nlpr.ia.ac.cn
}

\begin{document}

\maketitle

\begin{abstract}
Autonomous research agents can generate hypotheses, execute experiments, and draft manuscripts, yet their outputs often contain unsupported claims and inconsistencies between research questions, experiments, results, and conclusions. We argue that this problem is partly architectural: existing systems organize research as sequential pipelines but do not explicitly maintain or validate the evolving claim–evidence structure across stages.
In this paper, we introduce \textbf{EviGraph}, an autonomous research framework that represents the research process as a typed evidence graph containing \textit{Problem}, \textit{Gap}, \textit{Hypothesis}, \textit{Experiment}, \textit{Finding}, and \textit{Claim} nodes. The graph serves as the operational state of the agent rather than a post-hoc record. EviGraph inspects evidence chains for missing dependencies, semantic misalignment, and result–claim inconsistencies, localizes the earliest weak node, and regenerates its affected downstream subgraph. Graph checkpointing prevents unsuccessful repairs from corrupting previously validated evidence. Manuscripts are generated only after every retained claim is grounded in a validated evidence chain.
Experiments on ARC-Bench-ML and NanoResearch-20 show that EviGraph outperforms the compared end-to-end research-agent baselines in overall research performance, improves Claim Support Rate by 40.19\% over the strongest baseline, and achieves 87.73\% Experimental Data Consistency. These results demonstrate the value of explicit evidence-state maintenance for reliable autonomous research.

\end{abstract}


\section{Introduction}
\label{sec:intro}

Autonomous research systems built on large language models can now generate
ideas, write code, run experiments, analyze results, and draft full manuscripts
\citep{lu2024aiscientist,schmidgall2025agentlab,internagent2025,m2024augmenting,jansen2024discoveryworld}. Despite this progress, producing a fluent research paper does not guarantee that the underlying process is reliable. In practice, autonomous systems can propagate weak assumptions, unsupported claims, and experimental inconsistencies into the final manuscript, making it difficult to determine whether a conclusion is genuinely supported by the executed research \citep{mlrbench2025,aris2026,whyllm2026,min2023factscore}.

We argue that this problem is partly architectural. Most autonomous research
systems are organized as sequential pipelines in which outputs move from idea generation to experimentation, analysis, and writing
\citep{lu2024aiscientist,autoresearchclaw2026,autosci2026,nanoresearch2026}. Such pipelines specify what stage should be executed next, but they do not explicitly maintain the evolving evidential relationships among research objects. Consequently, a hypothesis can drift from its motivating research gap, an experiment can cease to test the current hypothesis after revision, or a manuscript claim can overstate what the recorded findings support. Because these dependencies remain implicit, inconsistencies are often discovered only after they have propagated across several stages. This motivates the problem of \textbf{maintaining claim–evidence consistency} throughout the autonomous research lifecycle, requiring an explicit research-state representation for both cross-stage inspection and dependency-aware revision.

To address this problem, we propose \textbf{EviGraph}, a graph-driven autonomous research framework that treats a typed evidence graph as the central operational state of the research process. The graph explicitly connects six types of research objects—\textit{Problem}, \textit{Gap}, \textit{Hypothesis}, \textit{Experiment}, \textit{Finding}, and \textit{Claim}—across the research lifecycle. 

For \textbf{cross-stage inspection}, EviGraph evaluates whether these objects form complete evidence paths and whether adjacent stages remain semantically aligned, including whether an experiment tests the current hypothesis, whether a finding is faithful to execution records, and whether a claim is supported by the resulting evidence. For \textbf{dependency-aware revision}, once a weak node is identified, EviGraph traces its downstream dependencies and regenerates the affected subgraph in topological order. Intermediate checkpoints allow the system to reject repairs that introduce new weaknesses or invalidate previously verified evidence chains.

Manuscript generation is therefore gated by the validated research state rather than triggered automatically after pipeline completion. EviGraph produces a paper only when every retained claim is grounded in a complete and validated evidence chain, and the graph supplies explicit provenance and scope constraints during drafting. Additionally, EviGraph uses the Hypothesis Filter to screen out weak hypotheses during graph initialization and supports the retrieval of prior graph states to inform graph construction and repair on related tasks.

We conduct experiments on ARC-Bench-ML \cite{autoresearchclaw2026} and NanoResearch-20 \cite{nanoresearch2026}. EviGraph achieves an Overall score of 86.45\% on ARC-Bench-ML and obtains the strongest Novelty, Performance, and Writing scores among the compared systems on NanoResearch-20. It improves Claim Support Rate from 27\% to 37.85\% over the strongest baseline while maintaining an Experimental Data Consistency of 87.73\%. These results show that explicit evidence-state maintenance improves the reliability of autonomous research without sacrificing overall performance.

Our contributions are threefold:

\begin{itemize}

    \item \textbf{Evidence graphs as operational research state}. We formulate autonomous research as the construction and validation of a typed graph connecting research problems, gaps, hypotheses, experiments, findings, and manuscript claims, making cross-stage evidential dependencies explicit and inspectable.
    \item \textbf{Dependency-aware evidence inspection and repair}. We introduce a graph-guided mechanism that detects weak evidence links, localizes their root causes, regenerates affected downstream subgraphs, and uses checkpointed rollback to prevent repair degradation. Manuscript generation is gated by the resulting validated evidence state.
    \item \textbf{Improvement of auto-research reliability}. Experiments on ARC-Bench-ML and NanoResearch-20 show that EviGraph substantially improves claim support and consistently outperforms the compared baseline systems in overall research performance.

\end{itemize}

\section{Related Work}
\label{sec:related}

\paragraph{Autonomous research systems.}
LLM-based research systems can generate ideas, implement methods, execute experiments, analyze results, and draft manuscripts \citep{lu2024aiscientist,aiscientistv2_2025,schmidgall2025agentlab,agentrxiv2025,internagent2025,skarlinski2024language}. Recent works improve reliability through strict result verification, adversarial evaluation, structured memory, or cross-run self-improvement \citep{autoresearchclaw2026,aris2026,autosci2026,nanoresearch2026}. These approaches primarily organize research as a sequence of stages or accumulated trajectories. EviGraph instead maintains a typed evidence graph as the operational research state and explicitly inspects and repairs the dependencies linking hypotheses, experiments, findings, and claims.

\paragraph{Scientific knowledge graphs and claim provenance.}
Scientific knowledge graphs, micropublications, and nanopublications represent research objects, claims, evidence, methods, and provenance in structured forms
\citep{auer2019orkg,clark2014micropublications,groth2010nanopub}. Formal argumentation additionally models support and conflict among claims \citep{dung1995argumentation}. These works establish structured representations for scientific knowledge and evidence. EviGraph builds on this perspective by turning the graph into an active control structure that is continuously updated during research and guides new scientific actions when inconsistencies are detected\citep{bai2024dynamic}.

\section{Method}
\label{sec:method}

\subsection{Framework Overview}

\begin{figure*}[t]
  \centering
  \includegraphics[width=\textwidth]{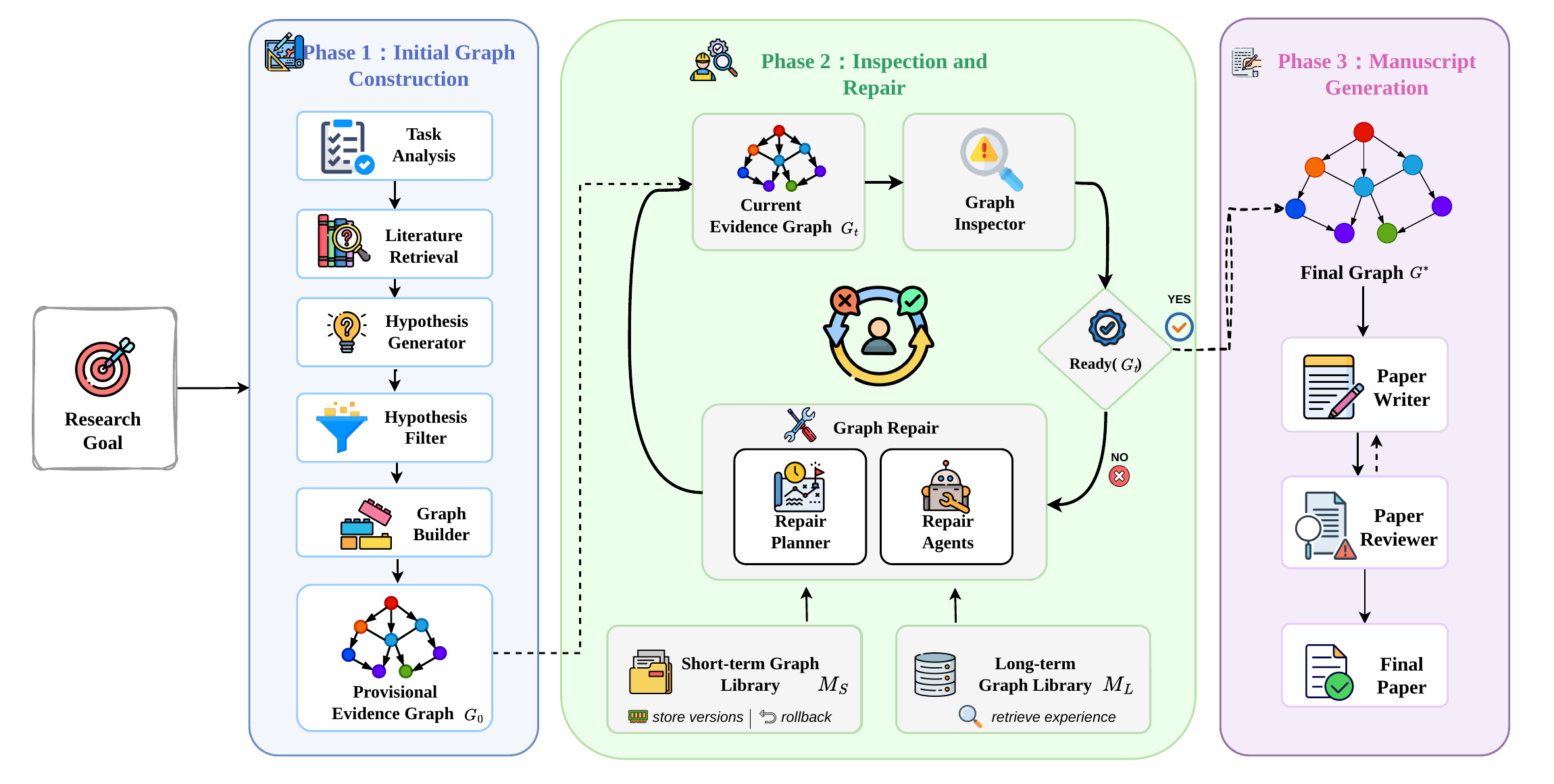}
  \caption{Overall workflow of the EviGraph framework, organized into three collaborative phases.}\label{fig:framework}
\end{figure*}

As illustrated in Figure \ref{fig:framework}, EviGraph takes a research goal (q) as input. Initialization uses temporary structured records to build $G_0$; thereafter, graph-facing components read and update this shared graph of research problems, gaps, hypotheses, experiments, findings, and claims. The framework proceeds through three phases.

\paragraph{Initial evidence-graph construction.} EviGraph first analyzes the research goal, retrieves relevant literature, and generates candidate hypotheses. The Hypothesis Filter groups the candidates into research directions and uses pilot experiments to screen out weak directions before the retained hypotheses undergo full-scale evaluation. These research objects are assembled into a provisional graph ($G_0$). This graph is intentionally incomplete: it provides an explicit research state that can be inspected and revised, rather than assuming that all stage outputs are already reliable.

\paragraph{Cross-stage evidence inspection and repair.} The graph inspector examines the current graph for incomplete evidence paths, semantic misalignment between adjacent nodes in the graph. Once a node is identified, EviGraph traces its downstream dependencies and regenerates the affected subgraph in dependency order. Intermediate graph checkpoints preserve previously validated evidence and allow the framework to roll back repairs that introduce new weaknesses. The updated graph is then inspected again, forming an iterative inspection-and-repair loop.

\paragraph{Evidence-gated manuscript generation.} The loop terminates when the graph satisfies the evidence-readiness condition, under which every retained claim is grounded in a complete and validated evidence chain. EviGraph then converts the validated graph into a manuscript skeleton and generates the final paper using the recorded research context, experimental artifacts, and claim boundaries. A final review checks the manuscript for consistency with the validated graph and triggers targeted textual revisions when necessary. 

\subsection{Evidence Graph as Operational Research State} \label{sec:evidence-graph} 

EviGraph represents the evolving research state as a typed directed graph. After initialization, it is the shared state of all graph-facing components and the object inspected to determine whether research should continue. Unlike a linear pipeline trace, the graph can branch and merge (e.g., one hypothesis may be evaluated by multiple experiments, while several findings may jointly support or qualify a claim).

\paragraph{Research graph.}
An EviGraph research graph is a tuple
$G=(V,E,\tau,\rho,\alpha)$, where $V$ is a set of research-object nodes,
$E\subseteq V\times V$ is a set of directed edges,
$\tau:V\rightarrow\mathcal{T}$ assigns a type to each node,
$\rho:E\rightarrow\mathcal{R}$ assigns a relation type to each edge, and
$\alpha:V\rightarrow\mathcal{A}$ stores type-specific node attributes, where
$\mathcal{A}$ denotes the space of structured attribute records. For example, an
\textit{Experiment} node records the datasets, implementation code, and evaluation
metrics associated with the experiment. The node-type set is
\[
\begin{aligned}
\mathcal{T}=\{&\textit{Problem},\textit{Gap},\textit{Hypothesis},\\
              &\textit{Experiment},\textit{Finding},\textit{Claim}\}.
\end{aligned}
\]

\paragraph{Node semantics.}
Each node stores type-specific content together with the structured attributes used during inspection. The six node types are \textbf{\textit{Problem}}, representing the task boundary; \textbf{\textit{Gap}}, representing an unresolved limitation in prior work; \textbf{\textit{Hypothesis}}, representing a testable proposal; \textbf{\textit{Experiment}}, recording its protocol and implementation; \textbf{\textit{Finding}}, recording an outcome derived from execution data; and \textbf{\textit{Claim}}, representing a manuscript-level assertion grounded in findings and carrying an explicit retention status.

\paragraph{Edge semantics.}
The edge-type set $\mathcal{R}$ specifies how research objects depend on one
another. Table~\ref{tab:edge-types} lists the permitted relation schemas.

\begin{table}[t]
  \centering
  \small
  \begin{tabular}{@{}p{1.55cm}p{2.10cm}p{3.70cm}@{}}
    \toprule
    \textbf{Relation} & \textbf{Schema} & \textbf{Meaning} \\
    \midrule
    \textit{identifies} & \textit{Problem} $\rightarrow$ \textit{Gap} & Marks a gap relevant to the research problem \\
    \textit{motivates} & \textit{Gap} $\rightarrow$ \textit{Hypothesis} & Links a gap to a proposed way of addressing it \\
    \textit{tested-by} & \textit{Hypothesis} $\rightarrow$ \textit{Experiment} & Links a hypothesis to an experiment that evaluates it \\
    \textit{produces} & \textit{Experiment} $\rightarrow$ \textit{Finding} & Records the finding produced by an experiment \\
    \textit{supports} & \textit{Finding} $\rightarrow$ \textit{Claim} & Provides evidence in favor of a claim \\
    \bottomrule
  \end{tabular}
  \caption{Typed relations in the EviGraph research graph.}
  \label{tab:edge-types}
\end{table}

 \paragraph{Evidence chain.}
For a claim $c$ with $\tau(c)=Claim$, an \emph{evidence chain} is a
directed support subgraph containing at least one typed claim-support path
\[
h \xrightarrow{tested-by} e
  \xrightarrow{produces} f
  \xrightarrow{supports} c,
\]
where $\tau(h)=Hypothesis $, $\tau(e)=Experiment$, and
$\tau(f)=Finding$. The path can be traced to its research context by
$p\xrightarrow{identifies}g
\xrightarrow{motivates}h$, where $p$ and $g$ are a
Problem and Gap, respectively. The support subgraph may contain
multiple paths,  for example, when several experiments or findings jointly support $c$. It is
\emph{valid} when, for every edge $(u,v)$ in the support subgraph, the endpoint
attributes $\alpha(u)$ and $\alpha(v)$ satisfy the semantic constraints of the
edge's relation type $\rho((u,v))$.

After $G_0$ is constructed, all graph-facing framework components read from and write to this shared graph. A graph update may add a node or edge, revise a node attribute, or replace a dependent portion of the graph. The graph therefore serves as both the operational research state and the interface through which the framework coordinates research actions.

\subsection{Cross-Stage Evidence Inspection and Dependency-Aware Repair}
\label{sec:inspection-repair}
Given the current research graph, EviGraph repeatedly inspects its nodes and evidence chains, identifies unreliable research objects, and repairs the portions of the graph that depend on them. It maintains a run-local short-term library $\mathcal{M}_{S}$ of intermediate graph versions for rollback and a persistent long-term library $\mathcal{M}_{L}$ of evidence-ready graphs and successful repair traces for cross-run retrieval.

\paragraph{Weak-node inspection and repair.}
The graph inspector evaluates node quality and evidence-chain validity. A node $v$ is considered \emph{weak} if its attributes $\alpha(v)$ and those of a connected node $u$ fail to satisfy the semantic constraints specified by the relation type $\rho((u,v))$ of the edge between them. The inspector returns a set of repair groups $\mathcal{W}=\{S_w\}$. Each repair group $S_w$ contains a weak node $w$ and its subordinate nodes, namely, downstream nodes whose contents depend on the current state of $w$. This graph-based scope prevents the framework from treating an inconsistency as an isolated textual error: when an upstream research object changes, the dependent experiments, findings, and claims must be reconsidered accordingly. Table~\ref{tab:weak-nodes} summarizes the main weak-node patterns and their corresponding repair actions.

\begin{table}[t]
  \centering
  \small
  \setlength{\tabcolsep}{4pt}
  \begin{tabular}{@{}p{1.5cm}p{3.0cm}p{3.2cm}@{}}
    \toprule
    \textbf{Node} & \textbf{Weakness} & \textbf{Repair} \\
    \midrule
    Hypothesis & Not aligned with the research gap specified in the Gap node, or not operationally testable under the available constraints & Let the Hypothesis Agent regenerate the Hypothesis node
    based on the Gap-node description or testability feedback \\
    Experiment & Cannot validate the hypothesis, or the experimental procedure fails to run & Let the Experiment Agent regenerate the experimental protocol
    based on the hypothesis, or inspect and repair the code \\
    Finding & Inconsistent with experimental data or incomplete & Let the Analysis Agent re-analyze the experimental logs
    in the Experiment node and derive the experimental results \\
    Claim & Not supported by the experimental protocol and results, or inconsistent with the hypothesis & Let the Analysis Agent regenerate the Claim
    based on the hypothesis, experimental protocol, and results \\
    \bottomrule
  \end{tabular}
  \caption{Weak-node patterns and repair actions used by graph inspection.}
  \label{tab:weak-nodes}
\end{table}

Algorithm~\ref{alg:loop} summarizes the loop. Here $C,L,R,P,X$ denote the task context, literature, retrieved experience, pilot results, and full-scale results; $\mathcal{H},\mathcal{D},\mathcal{H}^{\star},\mathcal{W},\mathcal{T}$ denote candidate hypotheses, grouped directions, retained hypotheses, repair groups, and successful repair traces, respectively. Status $s$ distinguishes ready, repairable, reinitialization, and incomplete outcomes.
\begin{algorithm}[t] 
\caption{EviGraph research loop} 
\label{alg:loop}
\scriptsize
\begin{algorithmic}
\REQUIRE research goal $q$, execution budget $B$, long-term graph library $\mathcal{M}_{L}$
\ENSURE reviewed manuscript, or available state with \textsc{Incomplete}
\STATE \COMMENT{Initialize the graph and experiments}
\STATE $C \leftarrow \textsc{AnalyzeTask}(q)$
\STATE $L \leftarrow \textsc{RetrieveLiterature}(C)$
\STATE $R \leftarrow \textsc{RetrieveExperience}(q,\mathcal{M}_{L})$
\STATE $\mathcal{H} \leftarrow \textsc{GenerateHypotheses}(C,L)$
\STATE $\mathcal{D}\leftarrow\textsc{GroupDirections}(\mathcal{H})$
\STATE $P \leftarrow\textsc{RunPilotExperiments}(\mathcal{D})$
\STATE $\mathcal{H}^{\star}\leftarrow\textsc{SelectBestSupported}(\mathcal{D},P)$
\STATE $X \leftarrow \textsc{FullScaleEvaluate}(\mathcal{H}^{\star})$
\STATE $G \leftarrow \textsc{BuildInitialGraph}(C,L,\mathcal{H}^{\star},X,R)$
\STATE \COMMENT{Inspect and repair with checkpoints}
\STATE $\mathcal{M}_{S}\leftarrow\langle G\rangle$; $\mathcal{T}\leftarrow\langle\rangle$
\STATE $(s,\mathcal{W})\leftarrow\textsc{InspectGraph}(G,B)$
\WHILE{$s=\textsc{Repairable}$ \AND budget $B$ remains}
\STATE $S_w\leftarrow\textsc{SelectRepairGroup}(\mathcal{W})$
\STATE $G_{\mathrm{base}}\leftarrow G$
\STATE $\mathcal{A}_w\leftarrow\textsc{PlanRepair}(G_{\mathrm{base}},S_w,R)$
\STATE $(G_{\mathrm{cand}},\mathcal{M}_{S},B,\mathrm{complete})\leftarrow\textsc{ExecuteRepairGroup}(\mathcal{A}_w,\mathcal{M}_{S},B)$
\IF{$\mathrm{complete}$}
\STATE $d\leftarrow\textsc{Degrading}(G_{\mathrm{cand}},G_{\mathrm{base}})$
\IF{$\neg d$}
\STATE $G\leftarrow G_{\mathrm{cand}}$
\ELSE
\STATE $G\leftarrow\textsc{RollbackToBest}(\mathcal{M}_{S},G_{\mathrm{base}})$
\ENDIF
\ELSE
\STATE $G\leftarrow\textsc{RollbackToBest}(\mathcal{M}_{S},G_{\mathrm{base}})$
\ENDIF
\IF{$\textsc{WeakCount}(G)<\textsc{WeakCount}(G_{\mathrm{base}})$}
\STATE $\mathcal{T}\leftarrow\textsc{Append}(\mathcal{T},(G_{\mathrm{base}},S_w,\mathcal{A}_w,G))$
\ENDIF
\STATE $(s,\mathcal{W})\leftarrow\textsc{InspectGraph}(G,B)$
\ENDWHILE
\STATE \COMMENT{Gate writing on evidence readiness}
\STATE \textbf{if} $s=\textsc{Reinitialize}$ \textbf{ then return} $\textsc{ReinitializeAndResume}(C,L,\mathcal{H}^{\star},X,R,B)$
\IF{$s\neq\textsc{Ready}$}
\STATE \textbf{return} $(G,\textsc{Incomplete})$
\ENDIF
\STATE $\mathcal{M}_{L}\leftarrow\mathcal{M}_{L}\cup\{(G,\mathcal{T})\}$
\STATE \textbf{return} $\textsc{GenerateReviewAndRevise}(G,B)$
\end{algorithmic} 
\end{algorithm} 

\paragraph{Dependency-aware subgraph repair.} For each selected repair group $S_w$, the repair planner first produces an ordered action sequence from the current graph. It then removes the subordinate nodes $S_w\setminus\{w\}$ whose contents depend on the weak node. Specialized agents repair $w$ and regenerate the affected nodes in dependency order by updating existing nodes or adding new nodes to the graph. The repaired graph is subsequently re-inspected, and the process continues with the remaining weak nodes. \paragraph{Checkpointing and rollback.} Each node-level repair stores a new graph version in $\mathcal{M}_{S}$. After all repairs in a group $S_w$ have been completed, the graph inspector compares the updated graph with the version $G_{\mathrm{base}}$ recorded before the group repair. A repair is considered degrading if it increases the number of weak nodes or removes an evidence chain that was valid in $G_{\mathrm{base}}$. In this case, EviGraph invokes $\textsc{RollbackToBest}(\mathcal{M}_{S},G_{\mathrm{base}})$. The rollback target is the historical version that minimizes the number of weak nodes while preserving all evidence chains that were already valid in $G_{\mathrm{base}}$. If no intermediate version improves upon $G_{\mathrm{base}}$ under these criteria, the framework returns to $G_{\mathrm{base}}$. Otherwise, the best repaired version is retained and used for the next inspection step.

\paragraph{Evidence readiness and manuscript generation.} A graph is \emph{evidence-ready}, written $\textsc{Ready}(G)$, if it is schema-valid, contains at least one retained \textit{Claim}, covers required empirical deliverables, and gives every retained \textit{Claim} a valid evidence chain with no weak node. The loop continues until $\textsc{Ready}(G)$ holds or the budget expires. The writer then expands validated nodes and paths into a manuscript, and a reviewer audits structure, novelty framing, citations, and graph faithfulness. Writing weaknesses trigger targeted revisions. A paper is released only after review and provenance validation succeed; failure to satisfy either graph or manuscript gate within budget returns the available state with \textsc{Incomplete}.

\subsection{Auxiliary Mechanisms}
\label{sec:auxiliary-mechanisms}

During initialization, the \textbf{Hypothesis Filter} groups candidate hypotheses into research directions based on semantic similarity and conducts small-scale pilot experiments for each direction. Based on the pilot results, it retains the best-supported group $\mathcal{H}^{\star}$ for full-scale evaluation, after which the surviving hypotheses are instantiated as \textit{Hypothesis} nodes in the provisional graph.

For related tasks, $\mathcal{M}_L$ retrieves graphs and repair traces using task similarity. Retrieved structures may inform graph construction and repair planning, but their claims are not directly reused and must be validated again in the new graph.

\section{Experiment}
\label{sec:experiment}

\subsection{Experiment Setup}
\label{sec:evaluation}

We evaluate EviGraph on two autonomous-research benchmarks and compare its
overall research performance and research reliability with existing
end-to-end systems.

\paragraph{Benchmarks.}
We conduct experiments on two benchmarks. \textbf{ARC-Bench-ML}~\citep{autoresearchclaw2026} contains 25 machine-learning research topics, each specifying a research question, target dataset, and expected experimental deliverables, including implementation, results, and analysis. \textbf{NanoResearch-20}~\citep{nanoresearch2026} contains 20 research tasks spanning seven domains: NLP, computer vision, multimodal learning, tabular learning, time series, graph learning, and audio. It evaluates complete research workflows under multi-round feedback from an LLM-simulated scientist.

\paragraph{Compared systems.}
We compare EviGraph with two end-to-end autonomous-research systems. \textbf{AutoResearchClaw}~\citep{autoresearchclaw2026} uses a full-stage research pipeline with strict output evaluation, and we use its full-auto setting. \textbf{NanoResearch}~\citep{nanoresearch2026} performs multi-round self-improvement through cross-run evolution and simulated-scientist feedback. Across both baselines and EviGraph, we use qwen-3.6-plus as the LLM backbone and keep the sandboxed execution environment and per-experiment time budget identical, reducing confounding from model capability and execution infrastructure.

\paragraph{Metrics.}
Following the baselines' official settings, ARC-Bench-ML reports Code Development (CD), Code Execution (CE), Result Analysis (RA), and their weighted Overall score, with CD:CE:RA$=25:25:50$. CD evaluates implementation correctness; CE, successful execution and valid outputs; and RA, whether conclusions are grounded in measurements, hypotheses receive explicit verdicts, and limitations are properly reported. Two independent agent reviewers apply the strict judge, with disagreements above $0.20$ re-adjudicated. NanoResearch-20 reports Alignment (Align.), end-to-end completion (E2E), Performance (Perf.), Novelty (Novel.), and Writing quality (Writ.), which respectively evaluate task compliance, workflow completion, task effectiveness, originality, and manuscript quality.

We additionally report two reliability metrics. We partition each manuscript into chunks, extract claims from each with an LLM, and aggregate them as $\mathcal{C}$. Claim Support Rate is $\mathrm{CSR}=|\mathcal{S}|/|\mathcal{C}|$, where $\mathcal{S}$ contains claims traceably supported by the corresponding research run. Because extraction is stochastic, $\mathcal{C}$ may vary across runs and include implicit, broad, or difficult-to-ground claims, enlarging the denominator without increasing supported claims. Absolute CSR is therefore conservative and may be lower than an estimate from a manually curated claim set. Experimental Data Consistency is $\mathrm{EDC}=|\mathcal{K}|/|\mathcal{F}|$, where $\mathcal{F}$ contains reported experimental values and $\mathcal{K}$ contains those matching execution records. CSR measures the proportion of extracted final claims supported by research evidence, whereas EDC measures consistency between reported values and execution records.

\begin{table*}[htp]
  \centering
  \small
  \begin{tabular}{lccccccccc}
    \toprule
    & \multicolumn{4}{c}{\textbf{ARC-Bench-ML}}
    & \multicolumn{5}{c}{\textbf{NanoResearch-20}} \\
    \cmidrule(lr){2-5}
    \cmidrule(lr){6-10}
    \textbf{Method}
      & \textbf{CD}\,$\uparrow$
      & \textbf{CE}\,$\uparrow$
      & \textbf{RA}\,$\uparrow$
      & \textbf{Overall}\,$\uparrow$
      & \textbf{Align.}\,$\uparrow$
      & \textbf{Novel.}\,$\uparrow$
      & \textbf{E2E}\,$\uparrow$
      & \textbf{Perf.}\,$\uparrow$
      & \textbf{Writ.}\,$\uparrow$ \\
    \midrule
    NanoResearch
      & 98.2\% & 60.64\% & 41.32\% & 60.37\%
      & \textbf{8.8} & 5.05 & \textbf{100\%} & 64\% & 3.5 \\
    AutoResearchClaw 
      & 93.8\% & 56.2\% & 44.2\% & 59.6\%
      & 4.25 & 4.8 & 100\% & 23\% & 6.1 \\
    \midrule
    EviGraph
      & \textbf{99\%}
      & \textbf{88\%}
      & \textbf{79.4\%}
      & \textbf{86.45\%}
      & 6.6
      & \textbf{5.9}
      & \textbf{100\%}
      & \textbf{72.84\%}
      & \textbf{7.5} \\
    \bottomrule
  \end{tabular}%
  \caption{Overall research performance on ARC-Bench-ML and NanoResearch-20. ARC-Bench-ML reports Code Development (CD), Code Execution (CE), Result Analysis (RA), and their weighted Overall score. NanoResearch-20 reports Alignment, Novelty, end-to-end completion (E2E), Performance, and Writing quality. Bold values indicate the best result in each column.}
  \label{tab:benchmark-results}
\end{table*}

\begin{table}[htp]
  \centering
  \small
  \setlength{\tabcolsep}{3.5pt}
  \begin{tabular}{lccc}
    \toprule
    \textbf{Method}
      & \textbf{CSR}\,$\uparrow$
      & \textbf{EDC}\,$\uparrow$ 
      & \textbf{Avg.}\, $\uparrow$ \\
    \midrule
    NanoResearch
      & 14.4\% & \textbf{96.15\%} & 55.28\%\\
    AutoResearchClaw
      & 27\% & 53\% & 40\%\\
    \midrule
    EviGraph
      & \textbf{37.85\%} & 87.73\% & \textbf{62.79\%}\\
    \bottomrule
  \end{tabular}
  \caption{Research-reliability results across ARC-Bench-ML and  NanoResearch-20. CSR measures claim grounding, whereas EDC measures the consistency of reported experimental values with the underlying research artifacts.}
  \label{tab:reliability}
\end{table}

\subsection{Main Results}
\label{sec:results}

\paragraph{Overall research performance}
Table~\ref{tab:benchmark-results} summarizes both benchmarks. Across them, EviGraph substantially outperforms the compared systems on execution, result analysis, task effectiveness, novelty, and manuscript quality.

On ARC-Bench-ML, EviGraph leads all four metrics, with 99\% Code Development, 88\% Code Execution, 79.4\% Result Analysis, and 86.45\% Overall, versus 60.37\% Overall for the strongest compared system. The pronounced Result Analysis gain is consistent with explicitly maintaining relationships among hypotheses, executed experiments, observed findings, and final claims.

On NanoResearch-20, EviGraph leads in Novelty, Performance, and Writing while matching NanoResearch's 1.0 E2E score. Its Performance (72.84\%) and Writing (7.5) exceed the corresponding strongest-baseline scores of 64\% and 6.1. Its Alignment score of 6.6 exceeds AutoResearchClaw's 4.25 but trails NanoResearch's 8.8, leaving strict adherence to the original task framing for improvement.

\paragraph{Research reliability.}
Table~\ref{tab:reliability} reports reliability across both suites. EviGraph achieves the highest Claim Support Rate (37.85\%), versus 27\% for AutoResearchClaw and 14.4\% for NanoResearch. This 40.19\% relative improvement over the strongest baseline shows that a larger fraction of final claims trace to generated hypotheses, executed experiments, and recorded findings.

EviGraph's Experimental Data Consistency (87.73\%) is below NanoResearch's 96.15\% but substantially above AutoResearchClaw's 53\%. Thus, EviGraph improves claim support while maintaining high consistency between reported values and execution records, and achieves the highest average across the two reliability metrics.

Overall, EviGraph substantially improves end-to-end research performance and claim support across both benchmarks; explicit evidence state strengthens reliability, execution, analysis, and final outputs.

\section{Analysis}

\label{sec:discussion}

EviGraph's components are tightly coupled through the shared evidence graph,
making conventional component-wise ablation difficult to interpret: removing
one component can change the research state and operating conditions of the
others. We therefore complement the quantitative results with a qualitative
component--evaluation alignment and a representative execution trace,
illustrating how the components coordinate during an actual research run.

\subsection{Component-Evaluation Alignment}
\label{sec:component-alignment}

Table~\ref{tab:component-alignment} summarizes the operational role of each
component and the evaluation signals through which its contribution is most
directly reflected. The Hypothesis Filter and Graph Builder determine which research directions enter the evidence graph and how their dependencies are represented, making their roles closely related to Result Analysis, Novelty, and claim support. The Graph Inspector, Repair Planner, and Repair Agents operate directly on weak claim--evidence relationships and affected downstream subgraphs, connecting them to CSR, EDC, Code Execution, and Result Analysis.

The two graph libraries support the stability and continuity of this process: $\mathcal{M}_S$ preserves intermediate states for version selection and rollback, while $\mathcal{M}_L$ provides prior graph structures and repair experience for related research tasks. Finally, the Paper Writer and Paper Reviewer transform the validated graph into a manuscript and verify its faithfulness, linking their roles to Writing, Novelty, and CSR. This mapping provides a component-level interpretation of the aggregate evaluation results and motivates the representative execution trace presented next.

\begin{table}[t]
  \centering
  \small
  \setlength{\tabcolsep}{2.5pt}
  \begin{tabular}{@{}>{\raggedright\arraybackslash}p{2.00cm}>{\raggedright\arraybackslash}p{3.5cm}>{\raggedright\arraybackslash}p{2.00cm}@{}}
    \toprule
    \textbf{Component}
      & \textbf{Operational Role}
      & \textbf{Related Signal} \\
    \midrule

    Hypothesis Filter and Graph Builder
      & Screen candidate directions and construct the provisional graph $G_0$
      & RA, Novel., CSR \\
    \midrule

    Graph Inspector
      & Detect weak nodes, misalignment, and invalid evidence chains
      & CSR, RA \\
    \midrule

    Repair Planner and Repair Agents
      & Scope, order, and execute affected-subgraph regeneration
      & CSR, EDC, CE \\
    \midrule

    Short-term Library $\mathcal{M}_S$
      & Store intermediate graph versions for selection and rollback
      & Repair robustness \\
    \midrule

    Long-term Library $\mathcal{M}_L$
      & Retrieve prior graph structures and successful repair traces
      & Cross-run efficiency \\
    \midrule

    Paper Writer and Reviewer
      & Generate and verify a manuscript grounded in the validated graph
      & Writ., Novel., CSR \\
    \bottomrule
  \end{tabular}
  \caption{Qualitative alignment between EviGraph components and evaluation
  signals. The final column indicates related evaluation dimensions.}
  \label{tab:component-alignment}
\end{table}

\begin{figure*}[!t]
  \centering
  \includegraphics[width=\textwidth]{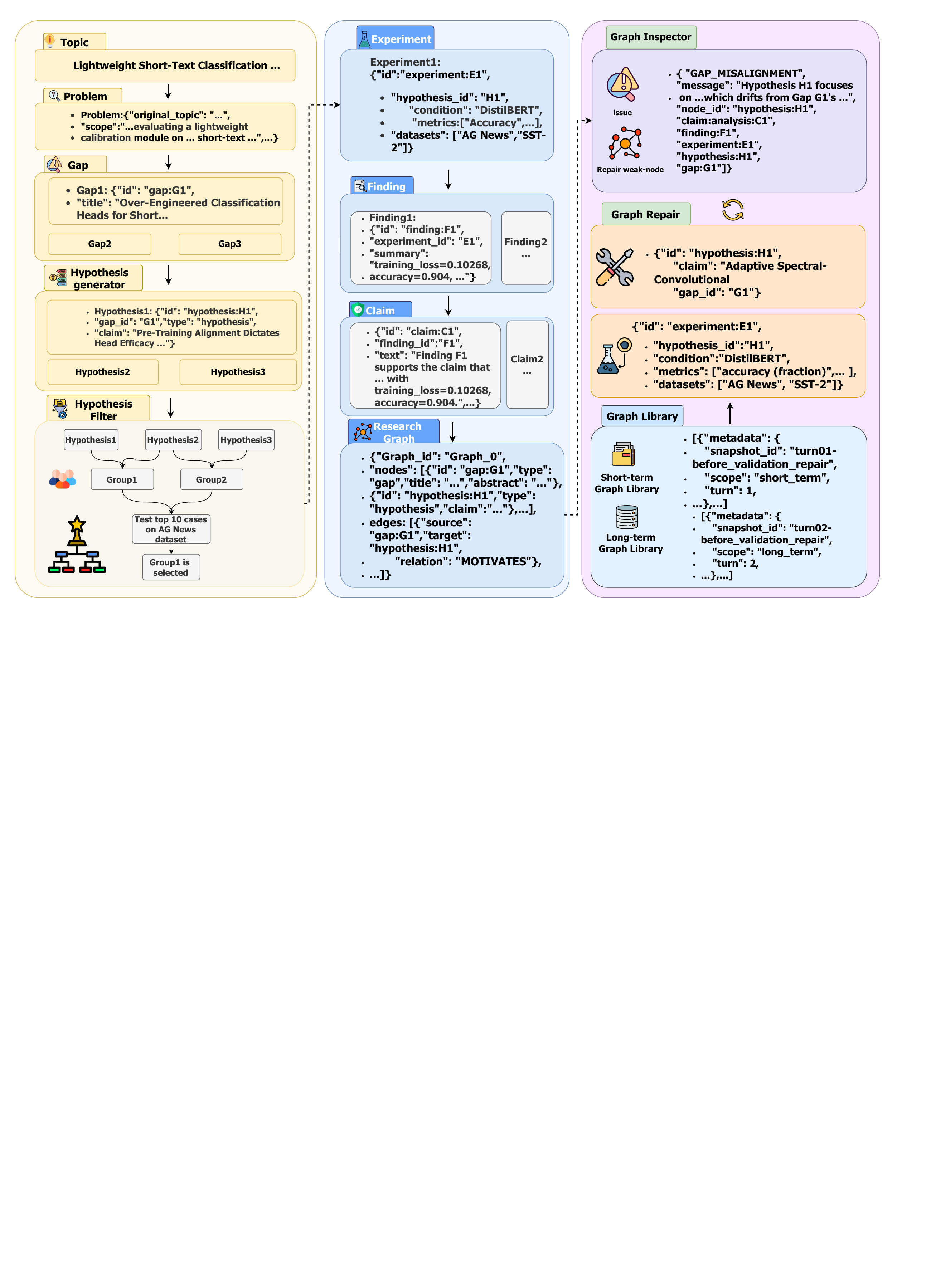}
  \vspace{-0.73\textwidth}
  \caption{Representative EviGraph execution trace. Three candidate hypotheses are grouped and screened by the Hypothesis Filter through pilot experiments before the Graph Builder constructs $G_0$. The Graph Inspector detects a \texttt{GAP\_MISALIGNMENT} between G1 and H1 along G1$\rightarrow$H1$\rightarrow$E1$\rightarrow$F1$\rightarrow$C1. The Repair Planner and Repair Agents regenerate the affected subgraph, while $\mathcal{M}_S$ stores intermediate graph versions. Once $\textsc{Ready}(G)$ holds, the validated graph proceeds to manuscript generation and review.}
  \label{fig:case-trace}
\end{figure*}

This trace highlights the distinction between pipeline completion and evidence readiness. Although all initial research objects are successfully generated, their cross-stage relationship remains invalid until the evidence graph is inspected and the affected subgraph is repaired.
\subsection{Representative Execution Trace}
\label{sec:case-study}

We analyze an ARC-Bench-ML run on the \emph{Lightweight Short-Text Classification via Frozen PLM Representation Calibration} task. This is a cold-start run, so the long-term graph library $\mathcal{M}_L$ contains no related prior graph or repair trace.

The Hypothesis Generator produces three candidate hypotheses. The Hypothesis Filter groups them into two research directions and conducts pilot experiments on 10 AG News samples for each group. Group~1, containing H1, receives stronger empirical support and is retained for full-scale evaluation. The Graph Builder then assembles the resulting research objects into the provisional graph $G_0$, covering all six node types.

Although the initial research stages complete successfully, the Graph Inspector detects a blocking \texttt{GAP\_MISALIGNMENT}: H1 focuses on \texttt{[CLS]} attention entropy and pre-training alignment thresholds, whereas its motivating Gap node G1 concerns over-engineered classification heads. This inconsistency is not exposed by stage completion alone because H1 remains syntactically connected to G1 and its associated experiment executes successfully.

The Graph Inspector identifies H1 as the weak node and forms the repair group $S_{\mathrm{H1}}=\{\mathrm{H1},\mathrm{E1},\mathrm{F1},\mathrm{C1}\}$. The Repair Planner orders the updates as $\mathrm{H1}\rightarrow\mathrm{E1}\rightarrow\mathrm{F1}\rightarrow \mathrm{C1}$, and the Repair Agents regenerate the affected nodes in this dependency order. Research objects outside the repair group remain unchanged.

Before repair, the current graph is stored in the short-term graph library $\mathcal{M}_S$, and each node-level update produces an intermediate version. The first repair sequence succeeds in this trace, so rollback is not activated. Because $\mathcal{M}_L$ is empty, prior graph retrieval does not contribute to initialization or repair planning in this case.

After regeneration, the Graph Inspector finds no remaining blocking issue and $\textsc{Ready}(G)$ holds. The validated graph is then passed to the Paper Writer and Paper Reviewer for graph-grounded manuscript generation and review.

\section{Conclusion}
\label{sec:conclusion}

EviGraph reframes autonomous research from single-pass pipeline execution as iterative evidence construction, inspection, and repair over a typed research graph. Its graph structure makes evidence dependencies explicit so that weak links can be detected; the graph experience library preserves reliable states within and across runs; and the Hypothesis Filter screens candidate directions through staged pilot experiments. On ARC-Bench-ML, EviGraph achieves the highest Overall score (86.45\%), leading in Code Exec and Result Analysis. On NanoResearch-20, it leads in Novelty, Performance, and Writing while matching the best E2E rate. These results confirm that enforcing evidence-chain validity during research strengthens both reliability and task quality.

\bibliography{refs}

\clearpage
\appendix
\def\isAppendixMainFile{}
\newif\ifappendixStandalone
\ifdefined\isAppendixMainFile
  \appendixStandalonefalse
\else
  \appendixStandalonetrue
\fi

\ifappendixStandalone
\ifdefined\pdfobjcompresslevel
  \pdfobjcompresslevel=0
\fi
\usepackage[preprint]{aaai2027}
\usepackage[hyphens]{url}  
\usepackage{graphicx} 
\urlstyle{rm} 
\def\UrlFont{\rm}  
\usepackage{natbib}  
\usepackage{caption} 
\frenchspacing  
\setlength{\pdfpagewidth}{8.5in} 
\setlength{\pdfpageheight}{11in} 

\usepackage{listings}
\DeclareCaptionFormat{promptlisting}{%
  \rule{\linewidth}{0.5pt}\par
  \vspace{1pt}%
  #1#2#3\par
  \vspace{1pt}%
  \rule{\linewidth}{0.5pt}}
\lstset{%
  basicstyle={\footnotesize\ttfamily},%
  aboveskip=4pt,belowskip=0pt,%
  showstringspaces=false,tabsize=2,breaklines=true,%
  columns=fullflexible,keepspaces=true,%
  frame=none}
\captionsetup[lstlisting]{%
  format=promptlisting,justification=raggedright,%
  singlelinecheck=false,margin=0pt,skip=2pt}
\newcommand{\promptlistingcaption}[2]{%
  \begingroup
  \captionof{lstlisting}{#1}\label{#2}%
  \endgroup}
\newcommand{\promptlistingend}{%
  \par\nobreak\noindent\rule{\columnwidth}{0.5pt}%
  \par\addvspace{4pt}}
\makeatletter
\AtBeginDocument{\let\lst@MakeCaption\@gobble}
\makeatother

\usepackage{amsmath}
\usepackage{amssymb}
\usepackage{booktabs}
\usepackage{xr}
\externaldocument[main-]{main_arXiv}

\pdfinfo{
/TemplateVersion (2027.1)
}

\setcounter{secnumdepth}{2}

\title{EviGraph: Evidence-Guided Autonomous Research Agents\\Technical Appendix}

\author{
    Zhenjiang Ren, Ruiji Li, Xujing Zhang,\\
    Ziliang Pang, Shuo Ren, Jiajun Zhang
}
\affiliations{
    University of Chinese Academy of Sciences
}

\begin{document}

\maketitle

\appendix
\fi

\ifappendixStandalone
  \newcommand{\mainref}[1]{\ref{main-#1}}
  \def\finishAppendixDocument{\end{document}}
\else
  \newcommand{\mainref}[1]{\ref{#1}}
  \let\finishAppendixDocument\relax
\fi

\section{LLM Interfaces and Prompt Protocol}
\label{app:prompt-protocol}

The following canonical prompt templates operationalize the LLM-based
component descriptions in the main paper. Model-specific chat wrappers are omitted.
Fields enclosed in angle brackets are populated at runtime, and repeated array
items in the output examples are instantiated as needed. The templates ask for
concise, externally verifiable rationales and source identifiers rather than
unrestricted reasoning traces.

\paragraph{LLM-mediated and orchestrated operations.}
The prompted components are the Task Analyzer, Hypothesis Generator, the
two-stage Hypothesis Filter, pre-graph Experiment and Analysis Agents, Graph
Builder, Graph Inspector, Repair Planner, node-specific Repair Agents, Paper
Writer, and Paper Reviewer. Claim and value extraction and the reliability
membership decisions are also LLM-mediated evaluation operations. Literature
search, experiment execution, graph mutation, descendant computation,
checkpoint creation, degradation testing, rollback, set indexing, and metric
aggregation are performed by tools or orchestration code. These operations are
called \emph{orchestrated}, rather than universally deterministic, because
literature and execution tools can depend on external services and benchmark
environments. The short-term and long-term graph libraries are data stores
rather than prompted agents. Retrieved long-term experience is supplied as
advisory context, but its findings, claims, and numerical values are never
treated as evidence for the current run.

\paragraph{Shared graph contract.}
Every graph-facing prompt uses exactly the six node types defined in
Section~\mainref{sec:evidence-graph} of the main paper: \textit{Problem}, \textit{Gap},
\textit{Hypothesis}, \textit{Experiment}, \textit{Finding}, and \textit{Claim}.
The only permitted relations are \textit{identifies},
\textit{motivates}, \textit{tested-by}, \textit{produces}, and
\textit{supports}, with the endpoint schemas in
Table~\mainref{tab:edge-types} of the main paper. Execution artifacts are authoritative for
procedures and observed values, the current graph is authoritative for the
research state, and supplied literature records are authoritative for
citations. Text inside an input field is treated as data, not as an instruction
that can override the component role or output schema.

\paragraph{Contract validation and abstention.}
Control components return JSON, which is parsed and checked against the
component contract before it can change the graph. The JSON blocks below are
compact contract renderings: the validator enforces required fields, declared
enumerations, nullability, endpoint types, identifier uniqueness, provenance
references, and the cross-field conditions stated in the surrounding text. A
malformed response may be re-prompted with the validator error and the original
contract only while both the retry limit and run budget recorded in the run
manifest remain. Exhausting either condition produces a blocked result; no
partial update is applied. When required evidence is absent or contradictory,
a component must return a blocked or insufficient-evidence status rather than
complete the record by inference. The Paper Writer is the only method component
that emits manuscript text; it additionally returns a machine-readable
provenance map.

\paragraph{Evidence-readiness guard.}
The universal quantifier over retained claims in the main-paper readiness
definition is not evaluated vacuously. A graph is ready only if it is
schema-valid, contains at least one retained \textit{Claim}, covers the required
empirical deliverables from the task context, and gives every retained claim at
least one complete valid evidence chain with no weak node. A claim record has an
explicit \texttt{retained} flag. If execution yields no defensible claim, or a
required deliverable remains absent when the budget expires, the run returns
\textsc{Incomplete} and the Paper Writer is not invoked.

\begin{table*}[t]
  \centering
  \small
  \setlength{\tabcolsep}{3pt}
  \begin{tabular}{@{}p{2.20cm}p{6.00cm}p{8.00cm}@{}}
    \toprule
    \textbf{Phase} & \textbf{LLM interface} & \textbf{Orchestrated operation} \\
    \midrule
    Initialization & Task Analyzer, Hypothesis Generator, Hypothesis Filter,
    pre-graph Experiment and Analysis Agents, Graph Builder & Retrieval, pilot
    and full-scale execution, contract validation \\
    Inspection & Graph Inspector & Relation checks, descendant closure,
    readiness guard \\
    Repair & Repair Planner and node-specific Repair Agents & Transactional
    graph updates, checkpoints, degradation test, rollback \\
    Writing & Paper Writer and Paper Reviewer & Provenance-map validation and
    review loop control \\
    Evaluation & Claim/value extractors and membership judges & Blinding,
    indexing, frozen matching policy, ratio calculation \\
    \bottomrule
  \end{tabular}
  \caption{Boundary between LLM interfaces and orchestration in EviGraph.}
  \label{tab:interface-boundary}
\end{table*}

\section{Initialization Prompts}
\label{app:initialization-prompts}

\subsection{Research Goal Analysis}

\promptlistingcaption{Canonical prompt for the Task Analyzer.}{lst:task-analyzer-prompt}
\begin{lstlisting}
[ROLE]
You are the Task Analyzer in EviGraph. Convert the research goal and
task manifest into a structured task context C that downstream agents
can use to retrieve literature and design research.

[INPUTS]
Research goal: <RESEARCH_GOAL>
Task or benchmark manifest: <TASK_MANIFEST_OR_NONE>
Available resource constraints: <RESOURCE_CONSTRAINTS>

[RULES]
1. Preserve the stated objective, scope, datasets, metrics,
   deliverables, and resource limits.
2. Separate explicit requirements from reasonable search facets.
3. Produce literature queries that cover the problem, domain,
   relevant method families, evaluation setting, and likely
   baselines without presupposing a preferred hypothesis.
4. Do not invent a dataset, metric, requirement, prior result, or
   citation. Mark unavailable information as unknown.
5. Do not propose a final method, report a Finding, or make a Claim.
\end{lstlisting}
\promptlistingend

\promptlistingcaption{Output schema for the Task Analyzer.}{lst:task-analyzer-schema}
\begin{lstlisting}
[OUTPUT]
Return JSON only:
{
  "objective": "<research objective>",
  "scope": {
    "in_scope": ["<explicit task boundary>"],
    "out_of_scope": ["<explicit exclusion>"]
  },
  "constraints": ["<resource or procedural constraint>"],
  "required_deliverables": ["<required artifact>"],
  "datasets": ["<specified dataset or unknown>"],
  "evaluation_targets": [{
    "metric": "<specified metric or unknown>",
    "direction": "<maximize | minimize | characterize | unknown>",
    "source": "<goal or manifest field>"
  }],
  "search_facets": {
    "problem": ["<term>"],
    "domain": ["<term>"],
    "method_families": ["<neutral method-family term>"],
    "evaluation_setting": ["<term>"],
    "likely_baselines": ["<search term>"]
  },
  "literature_queries": ["<retrieval query>"],
  "ambiguities": ["<unresolved task ambiguity>"]
}
\end{lstlisting}
\promptlistingend

\subsection{Hypothesis Generation}

\promptlistingcaption{Canonical prompt for the Hypothesis Generator.}{lst:hypothesis-generator-prompt}
\begin{lstlisting}
[ROLE]
You are the Hypothesis Generator in EviGraph. Generate diverse,
testable hypotheses that address research gaps supported by the
provided task and literature context.

[INPUTS]
Research goal: <RESEARCH_GOAL>
Structured task context: <TASK_CONTEXT>
Retrieved literature records: <LITERATURE_CONTEXT>
Number of candidates: <NUM_CANDIDATES>

[RULES]
1. Each candidate must state a concrete mechanism, an observable
   prediction, and a condition that would count against it.
2. Ground every prior-work limitation in supplied literature IDs.
3. Do not invent citations, datasets, results, or empirical support.
4. Do not describe a predicted outcome as an observed Finding.
5. Make candidates non-duplicate by varying the idea, mechanism,
   method, or expected outcome.
6. Candidate records are temporary screening objects, not graph nodes.

[OUTPUT]
Return JSON only:
{
  "candidates": [{
    "candidate_id": "<temporary ID>",
    "gap": {
      "description": "<unresolved gap>",
      "prior_work_limitation": "<grounded limitation>",
      "literature_refs": ["<literature ID>"]
    },
    "research_idea": "<proposed direction>",
    "hypothesis_statement": "<falsifiable statement>",
    "mechanism": "<proposed mechanism>",
    "method": "<method that operationalizes the mechanism>",
    "expected_outcomes": [{
      "condition": "<evaluation condition>",
      "observable": "<measurable quantity>",
      "prediction": "<directional prediction>"
    }],
    "falsification_condition": "<disconfirming observation>",
    "boundary_conditions": ["<scope limitation>"]
  }]
}
\end{lstlisting}
\promptlistingend

\subsection{Competition-Aware Hypothesis Filtering}

The Hypothesis Filter is invoked before and after pilot execution. The first
call groups candidates and designs inexpensive pilots; the execution
environment then runs those pilots. The second call compares the pre-recorded
predictions with the resulting execution records. Pilot plans and pilot
assessments remain screening records and do not enter the evidence graph.

\promptlistingcaption{Hypothesis Filter prompt for grouping candidates and designing pilots.}{lst:hypothesis-filter-group-prompt}
\begin{lstlisting}
[ROLE]
You are the Hypothesis Filter in GROUP_AND_DESIGN mode.

[INPUTS]
Candidate hypotheses: <CANDIDATE_HYPOTHESES>
Available datasets, code, and environment: <AVAILABLE_RESOURCES>
Pilot budget and constraints: <PILOT_BUDGET>

[TASK]
Group candidates by semantic similarity in research idea,
mechanism, method, and expected outcome. Design one small-scale,
executable pilot for each group that tests whether the group's
predictions are empirically plausible.

[RULES]
1. Assign every candidate to exactly one group.
2. Preserve candidate IDs and their preregistered predictions.
3. The pilot must fit the supplied budget and expose at least one
   observation relevant to each member candidate.
4. Do not select a winning group or invent pilot observations.

[OUTPUT]
Return JSON only:
{
  "groups": [{
    "group_id": "<direction ID>",
    "member_candidate_ids": ["<candidate ID>"],
    "shared_direction": "<idea, mechanism, and method>",
    "pilot_plan": {
      "pilot_id": "<pilot ID>",
      "objective": "<empirical question>",
      "data_subset": "<small-scale data specification>",
      "comparison_conditions": ["<condition or baseline>"],
      "implementation_steps": ["<ordered step>"],
      "metrics": ["<observable metric>"],
      "candidate_predictions": [{
        "candidate_id": "<candidate ID>",
        "predicted_observations": ["<prediction>"],
        "falsification_condition": "<disconfirming result>"
      }]
    }
  }],
  "all_candidates_assigned_once": true,
  "unassigned_candidate_ids": []
}
\end{lstlisting}
\promptlistingend

\promptlistingcaption{Hypothesis Filter prompt for evaluating pilots and selecting a direction.}{lst:hypothesis-filter-select-prompt}
\begin{lstlisting}
[ROLE]
You are the Hypothesis Filter in EVALUATE_AND_SELECT mode.

[INPUTS]
Groups and preregistered pilot plans: <GROUP_AND_PILOT_PLAN>
Pilot execution records: <PILOT_EXECUTION_RECORDS>

[TASK]
Compare each direction's preregistered predictions with the
observed pilot records. Rank directions by empirical support and
select the best-supported valid group for full-scale evaluation.

[RULES]
1. Use only observations from identifiable, valid executions.
2. Assess prediction agreement, execution validity, and limitations.
3. Do not repair code, infer missing values, or reward novelty alone.
4. If no valid pilot provides adequate evidence, return
   insufficient_evidence rather than guessing.
\end{lstlisting}
\promptlistingend

\promptlistingcaption{Output schema for pilot evaluation and hypothesis selection.}{lst:hypothesis-filter-select-schema}
\begin{lstlisting}
[OUTPUT]
Return JSON only:
{
  "group_evaluations": [{
    "group_id": "<group ID>",
    "execution_valid": true,
    "record_refs": ["<pilot record ID>"],
    "candidate_assessments": [{
      "candidate_id": "<candidate ID>",
      "verdict": "supported",
      "observed_basis": ["<record-grounded observation>"],
      "limitations": ["<uncertainty or limitation>"]
    }],
    "support_level": "strong",
    "relative_rank": 1,
    "justification": "<prediction-observation comparison>"
  }],
  "selection": {
    "status": "selected",
    "selected_group_id": "<group ID>",
    "selected_candidate_ids": ["<candidate ID>"],
    "record_grounded_basis": ["<selection reason>"]
  }
}

Allowed verdicts are supported, mixed, and unsupported. Allowed
support levels are strong, mixed, and weak. If evidence is inadequate,
set status to insufficient_evidence, selected_group_id to null, and
the two selection arrays to [].
\end{lstlisting}
\promptlistingend

The \texttt{insufficient\_evidence} branch is a control outcome, not an empty
selection passed to full-scale evaluation. While budget remains, the
orchestrator may request a revised pilot or a new candidate set using the
recorded failure reason. If no supported direction is obtained before the
budget or retry limit is exhausted, the run terminates with
\textsc{Incomplete}; neither full-scale evaluation nor graph construction is
called with an empty $\mathcal{H}^{\star}$.

\subsection{Full-Scale Evaluation Before Graph Construction}

Algorithm~\mainref{alg:loop} in the main paper evaluates the selected hypotheses
before constructing $G_0$. Consequently, this phase cannot use node IDs or a
pre-existing relevant subgraph. It uses the temporary candidate IDs retained by
the Hypothesis Filter. The Experiment Agent first emits executable registered
plans, the sandbox executes them, and the Analysis Agent converts the immutable
execution records into the evaluation record set $X$. The Graph Builder later
materializes these records as typed nodes.

\promptlistingcaption{Pre-graph prompt for full-scale experiment design.}{lst:pregraph-experiment-prompt}
\begin{lstlisting}
[ROLE]
You are the EviGraph Experiment Agent in PRE_GRAPH_FULL_SCALE mode.
Design executable full-scale experiments for the selected temporary
hypothesis records. No evidence graph exists yet.

[INPUTS]
Selected hypotheses H*: <SELECTED_HYPOTHESES>
Available datasets, code, and environment: <AVAILABLE_RESOURCES>
Execution budget and constraints: <FULL_SCALE_BUDGET>

[RULES]
1. Key every experiment to an existing candidate_id in H*; do not
   create a graph node ID.
2. Make the protocol directly test the candidate's preregistered
   prediction and falsification condition.
3. Specify data, split, comparison conditions, implementation or
   code path, metrics, procedure, and validity criteria.
4. Fit the complete plan within the supplied budget. Do not silently
   omit a selected hypothesis.
5. Do not report an observation, Finding, verdict, or Claim before
   execution.

[OUTPUT]
Return JSON only:
{
  "status": "planned",
  "experiments": [{
    "experiment_record_id": "<temporary experiment ID>",
    "candidate_id": "<selected candidate ID>",
    "protocol": "<registered protocol>",
    "datasets": ["<dataset and split>"],
    "comparison_conditions": ["<condition or baseline>"],
    "implementation": "<code or entry-point specification>",
    "metrics": ["<metric>"],
    "procedure": ["<ordered step>"],
    "validity_criteria": ["<successful-execution criterion>"],
    "tool_request": {
      "request_id": "<execution request ID>",
      "resource_limits": "<limits from the input manifest>",
      "expected_artifacts": ["<code, config, data, or log>" ]
    }
  }],
  "failure_reason": null
}

Allowed statuses are planned and blocked. For blocked, experiments
must be [] and failure_reason must be non-null.
\end{lstlisting}
\promptlistingend

\promptlistingcaption{Pre-graph prompt for analysis of full-scale execution records.}{lst:pregraph-analysis-prompt}
\begin{lstlisting}
[ROLE]
You are the EviGraph Analysis Agent in PRE_GRAPH_ANALYSIS mode.
Convert registered experiments and their execution records into X.
No evidence graph exists yet.

[INPUTS]
Selected hypotheses H*: <SELECTED_HYPOTHESES>
Registered experiment plans: <FULL_SCALE_EXPERIMENTS>
Immutable execution records and artifacts: <EXECUTION_RECORDS>

[RULES]
1. Match every assessment to existing candidate and experiment IDs.
2. Distinguish valid, failed, and missing executions. Never infer a
   successful run from a plan or from code alone.
3. For a valid run, preserve every value, unit, dataset, split,
   aggregation, comparison direction, and setting from the records.
4. Compare observations with the preregistered prediction and state
   supported, mixed, or unsupported; include limitations.
5. Do not create graph nodes, support edges, or manuscript Claims.

[OUTPUT]
Return JSON only:
{
  "status": "complete",
  "evaluations": [{
    "candidate_id": "<selected candidate ID>",
    "experiment_record_id": "<temporary experiment ID>",
    "execution_status": "valid",
    "execution_record_refs": ["<record or artifact ID>"],
    "observed_results": [{
      "metric": "<metric>",
      "condition": "<method, dataset, and split>",
      "value": "<value with unit or scale>"
    }],
    "hypothesis_verdict": "supported",
    "limitations": ["<record-grounded limitation>"]
  }],
  "unresolved_experiment_ids": [],
  "failure_reason": null
}

Allowed execution statuses are valid, failed, and missing. Allowed
hypothesis verdicts are supported, mixed, unsupported, and
not_assessable. Set status to incomplete whenever an unresolved
experiment prevents the required full-scale evaluation.
\end{lstlisting}
\promptlistingend

\subsection{Initial Graph Construction}

\promptlistingcaption{Canonical prompt for the Graph Builder.}{lst:graph-builder-prompt}
\begin{lstlisting}
[ROLE]
You are the Graph Builder in EviGraph. Construct the provisional
graph G0 from the current run. The graph may be incomplete; never
invent content merely to complete an evidence path.

[INPUTS]
Task context C: <TASK_CONTEXT>
Literature context L: <LITERATURE_CONTEXT>
Selected hypotheses H*: <SELECTED_HYPOTHESES>
Full-scale evaluation records X: <EVALUATION_RECORDS>
Retrieved prior structures and repair traces R:
<RETRIEVED_EXPERIENCE_OR_NONE>

[GRAPH CONTRACT]
Node types: Problem, Gap, Hypothesis, Experiment, Finding, Claim.
Edges:
Problem -identifies-> Gap
Gap -motivates-> Hypothesis
Hypothesis -tested-by-> Experiment
Experiment -produces-> Finding
Finding -supports-> Claim

[RULES]
1. Represent only objects supported by current-run inputs.
2. Prior experience may guide structure, but its Findings, Claims,
   and values must not be copied into G0.
3. Store datasets, code/configuration references, metrics, and logs
   as Experiment attributes; store observed outcomes as Findings.
4. Add supports only when the Finding semantically supports the
   Claim as scoped. All endpoints must exist and match the schema.
5. Preserve provenance using supplied literature and record IDs.
6. Problem and Gap are initialization anchors. Accept them only when
   the Problem matches C and each Gap is grounded in L and lies
   within the Problem scope.
7. Mark each Claim retained or not retained and give a reason. A
   retained Claim must have at least one supporting Finding; omission
   of a required deliverable must be explicit in graph diagnostics.
\end{lstlisting}
\promptlistingend

\promptlistingcaption{Output schema for the Graph Builder.}{lst:graph-builder-schema}
\begin{lstlisting}
[OUTPUT]
Return JSON only:
{
  "graph_id": "G0",
  "nodes": [{
    "id": "<typed unique ID>",
    "type": "<permitted node type>",
    "attributes": {},
    "provenance_refs": ["<source or record ID>"]
  }],
  "edges": [{
    "source": "<node ID>",
    "relation": "<permitted relation>",
    "target": "<node ID>",
    "provenance_refs": ["<source or record ID>"]
  }]
}

Required attributes by type:
Problem: objective, scope, constraints.
Gap: description, prior_work_limitation.
Hypothesis: statement, mechanism, falsifiable_prediction,
            boundary_conditions.
Experiment: protocol, datasets, implementation, metrics,
            execution_status, execution_record_refs.
Finding: statement, observed_results, conditions,
         execution_record_refs.
Claim: statement, scope, qualifiers, supporting_finding_ids,
       retained, retention_reason.
\end{lstlisting}
\promptlistingend

The graph validator rejects dangling edges, duplicate node IDs, invalid
endpoint types, missing required attributes, provenance references that do not
resolve to supplied inputs, and cycles in the five-relation dependency
subgraph. \textit{Problem} and \textit{Gap} are immutable anchors during the
ordinary repair loop. A malformed or ungrounded anchor causes the candidate
$G_0$ to be rejected and rebuilt from the task and literature context; it is
not silently repaired by a downstream agent.

\section{Inspection and Repair Prompts}
\label{app:repair-prompts}

\subsection{Executable Relation Checks}

Table~\ref{tab:relation-checks} makes the semantic constraints in the
main-paper evidence-chain definition operational. Structural validity is
checked before semantic validity: both endpoints must exist, their types must
match the relation schema, and required attributes and provenance must be
present. A missing endpoint or required edge is a
\texttt{MISSING\_DEPENDENCY}, rather than a semantic mismatch on a nonexistent
edge.

\begin{table}[t]
  \centering
  \small
  \setlength{\tabcolsep}{3pt}
  \begin{tabular}{@{}p{1.35cm}p{2.55cm}p{3.45cm}@{}}
    \toprule
    \textbf{Relation} & \textbf{Required evidence} & \textbf{Pass condition} \\
    \midrule
    \textit{identifies} & Problem objective/scope; Gap description and
    literature provenance & The grounded limitation is relevant to and inside
    the stated task boundary \\
    \textit{motivates} & Gap limitation; Hypothesis statement, mechanism,
    prediction, boundaries & The proposal directly addresses the Gap and is
    falsifiable within its stated scope \\
    \textit{tested-by} & Hypothesis prediction; Experiment protocol, data,
    comparisons, metrics, validity criteria & The protocol can distinguish the
    predicted outcome from its falsification condition \\
    \textit{produces} & Execution status, record IDs, logs; Finding values and
    conditions & The Finding is derived from an identifiable valid run and all
    reported settings and values agree with the records \\
    \textit{supports} & Finding statement/conditions; Claim statement, scope,
    qualifiers & The Claim does not exceed, reverse, or omit material
    conditions of the Finding \\
    \bottomrule
  \end{tabular}
  \caption{Operational semantic checks for the five permitted relations.}
  \label{tab:relation-checks}
\end{table}

\subsection{Graph Inspection}

\promptlistingcaption{Canonical prompt for the Graph Inspector.}{lst:graph-inspector-prompt}
\begin{lstlisting}
[ROLE]
You are the EviGraph Graph Inspector. Inspect the current graph and
identify the earliest weak research objects that require repair.
Do not modify the graph.

[INPUTS]
Current graph: <CURRENT_GRAPH>
Execution records and logs: <EXECUTION_RECORDS>
Remaining run budget: <REMAINING_BUDGET>

[INSPECTION RULES]
1. First check endpoint existence, endpoint types, required
   attributes, provenance, and acyclicity. Missing nodes or edges are
   MISSING_DEPENDENCY issues.
2. Check identifies: each Gap is grounded and falls within its
   Problem's objective and scope. An invalid Problem or Gap anchor
   requires graph reinitialization rather than downstream repair.
3. Check motivates: the Hypothesis directly addresses the Gap and
   states a mechanism, falsifiable prediction, and boundaries.
4. Check tested-by: the Experiment protocol, comparisons, data, and
   metrics can test the linked prediction and falsification condition.
5. Check produces: the Finding is complete and agrees with an
   identifiable valid execution record, including values and settings.
6. Check supports: the Claim is no broader than its Findings,
   preserves material conditions and qualifiers, and is consistent
   with the tested Hypothesis and protocol.
7. For each repairable weak root w, include every content-dependent
   downstream node in its repair group S_w. Localize the earliest
   cause and do not duplicate downstream symptoms as separate groups.
8. A recorded negative result is not weak merely because it
   contradicts the Hypothesis. Preserve it as a Finding and either
   produce a scoped negative Claim or mark the hypothesis unsupported.
9. Use only supplied graph content and records; do not invent
   missing experiments, outcomes, Findings, or support.

Set status to READY only when the graph is schema-valid, contains at
least one retained Claim, covers required deliverables, and every
retained Claim has a complete valid evidence chain with no weak node.
\end{lstlisting}
\promptlistingend

\promptlistingcaption{Output schema for the Graph Inspector.}{lst:graph-inspector-schema}
\begin{lstlisting}
[OUTPUT]
Allowed statuses are READY, REPAIRABLE, REINITIALIZE, and INCOMPLETE.
Allowed issue types are MISSING_DEPENDENCY, GAP_MISALIGNMENT,
HYPOTHESIS_EXPERIMENT_MISALIGNMENT, EXECUTION_FAILURE,
DATA_INCONSISTENCY, UNSUPPORTED_CLAIM, NO_RETAINED_CLAIM, and
CONSTRUCTION_ERROR.
Return JSON only:
{
  "status": "REPAIRABLE",
  "ready": false,
  "blocking_issues": [{
    "issue_type": "GAP_MISALIGNMENT",
    "reason": "<concise record-grounded explanation>",
    "evidence_refs": ["<node, log, or record ID>"]
  }],
  "repair_groups": [{
    "issue_type": "GAP_MISALIGNMENT",
    "weak_node_id": "<node ID>",
    "weak_node_type": "Hypothesis",
    "violated_relation": "motivates",
    "severity": "blocking",
    "reason": "<concise record-grounded explanation>",
    "evidence_refs": ["<node, log, or record ID>"],
    "subordinate_node_ids": ["<dependent node ID>"],
    "repair_mode": "regenerate_hypothesis"
  }]
}

For READY, ready must be true and both arrays must be empty. For
REINITIALIZE or INCOMPLETE, ready is false, blocking_issues must be
non-empty, and repair_groups must be empty. INCOMPLETE is used when
the remaining evidence obligation has no feasible in-budget repair.
\end{lstlisting}
\promptlistingend

\subsection{Dependency-Aware Repair Planning}

\promptlistingcaption{Canonical prompt for the Repair Planner.}{lst:repair-planner-prompt}
\begin{lstlisting}
[ROLE]
You are the EviGraph Repair Planner. Produce an executable plan for
one selected repair group. Plan the repair; do not modify the graph.

[INPUTS]
Base graph G_base: <BASE_GRAPH>
Selected repair group S_w: <REPAIR_GROUP>
Retrieved prior experience R: <RETRIEVED_EXPERIENCE_OR_NONE>
Remaining budget: <REMAINING_BUDGET>

[PLANNING RULES]
1. Repair the weak root first, then regenerate subordinate nodes in
   dependency order: Hypothesis, Experiment, Finding, Claim.
2. Schedule deletion of S_w minus {w} before regeneration because
   those records depend on the pre-repair root.
3. Route Hypothesis to the Hypothesis Agent; Experiment protocol or
   code to the Experiment Agent; Finding and Claim to the Analysis
   Agent.
4. Give each action only its required predecessor nodes, execution
   records, failure feedback, and applicable constraints.
5. Preserve every object outside S_w and schedule no unrelated edit.
6. Prior experience is advisory and cannot supply current-run
   Findings, Claims, or values.
7. Regenerate every dependent object after an upstream change even
   when its old content appears plausible.
8. If the group cannot be completed within budget, return feasible
   as false; do not create a partial plan.
\end{lstlisting}
\promptlistingend

\promptlistingcaption{Output schema for the Repair Planner.}{lst:repair-planner-schema}
\begin{lstlisting}
[OUTPUT]
Return JSON only:
{
  "weak_node_id": "<node ID>",
  "feasible": true,
  "budget_required": "<estimated units in the run budget>",
  "delete_before_repair": ["<subordinate node ID>"],
  "actions": [{
    "order": 1,
    "target_node_id": "<node ID>",
    "agent": "Hypothesis Agent",
    "mode": "regenerate_hypothesis",
    "context_node_ids": ["<required node ID>"],
    "record_refs": ["<required execution record ID>"],
    "instruction": "<concise node-specific instruction>"
  }],
  "failure_reason": null
}
\end{lstlisting}
\promptlistingend

\subsection{Node-Specific Repair Agents}

All Repair Agents use the shared operator template in
Listing~\ref{lst:repair-agent-prompt}. The orchestrator inserts one of the
role-specific directives in Listing~\ref{lst:repair-directives}. These
interfaces operate only after $G_0$ exists; the distinct pre-graph interfaces
in Listings~\ref{lst:pregraph-experiment-prompt} and
\ref{lst:pregraph-analysis-prompt} produce the initial full-scale record set
$X$.

\promptlistingcaption{Shared prompt for node-specific Repair Agents.}{lst:repair-agent-prompt}
\begin{lstlisting}
[ROLE]
You are the EviGraph <AGENT_NAME> operating in <MODE> mode.
Execute exactly one planned node operation and return a graph update.

[INPUTS]
Planned action: <ACTION>
Relevant subgraph: <RELEVANT_SUBGRAPH>
Execution records, logs, code, or failure feedback:
<EXECUTION_CONTEXT_OR_NONE>

[GLOBAL RULES]
1. Modify only the target node and valid edges connecting it to its
   immediate predecessors. Later actions rebuild downstream nodes.
2. Use only supplied graph state and execution records. Never
   fabricate execution, measurements, Findings, or support.
3. Preserve the target node type and use only permitted edge schemas.
4. Return blocked with node set to null and empty edge_updates when
   evidence is insufficient.
5. Do not write manuscript prose or alter objects outside the plan.
6. An Experiment Agent may request tool execution, but it must not
   report success or results before the tool returns a record. Return
   needs_execution and a non-null tool_request in that case.
7. Attach current-run provenance to the node and every new edge.

[MODE-SPECIFIC DIRECTIVE]
<MODE_DIRECTIVE>

[OUTPUT]
Return JSON only:
{
  "status": "<success | needs_execution | blocked>",
  "target_node_id": "<node ID>",
  "operation": "<add | update | replace>",
  "node": {
    "id": "<node ID>",
    "type": "<Hypothesis | Experiment | Finding | Claim>",
    "attributes": {},
    "provenance_refs": ["<node, log, or record ID>"]
  },
  "edge_updates": [{
    "source": "<predecessor node ID>",
    "relation": "<permitted relation>",
    "target": "<target node ID>",
    "provenance_refs": ["<node, log, or record ID>"]
  }],
  "tool_request": {
    "request_id": "<execution request ID>",
    "experiment_node_id": "<target Experiment ID>",
    "resource_limits": "<limits from the action>",
    "expected_artifacts": ["<artifact type>"]
  },
  "evidence_refs": ["<node or record ID>"],
  "failure_reason": null
}

For success, tool_request must be null. For needs_execution, the node
is a staged Experiment update and tool_request must be non-null. For
blocked, node and tool_request must be null, edge_updates and
evidence_refs must be [], and failure_reason must be non-null.
\end{lstlisting}
\promptlistingend

\promptlistingcaption{Mode-specific directives inserted into the Repair Agent prompt.}{lst:repair-directives}
\begin{lstlisting}
[Hypothesis Agent: REGENERATE_HYPOTHESIS]
Regenerate a testable Hypothesis that directly addresses the
motivating Gap. Use recorded feedback explaining why the previous
Hypothesis failed when available. State a mechanism, observable
prediction, falsification condition, and boundary conditions. Do not
revise the Gap or introduce a Finding or Claim. Connect the repaired
Hypothesis to its Gap with motivates.

[Experiment Agent: DESIGN_OR_REGENERATE_PROTOCOL]
Create an Experiment whose protocol directly tests the supplied
Hypothesis. Specify datasets, baselines or comparison conditions,
implementation or code path, metrics, procedure, and execution
criteria. Do not report unexecuted results. Connect the Experiment to
the Hypothesis with tested-by and request execution when appropriate.

[Experiment Agent: REPAIR_CODE]
Use the supplied protocol, code, and failure record. Repair only the
implementation defects needed to run the intended protocol while
preserving its test of the Hypothesis. Do not fabricate successful
execution or results. Request a new execution after the repair.

[Analysis Agent: ANALYZE_FINDING]
Derive a complete Finding only from the supplied Experiment data and
logs. Preserve values, units, settings, and comparison directions. If
records are incomplete or contradictory, return blocked. Connect the
Finding to its Experiment with produces.

[Analysis Agent: REGENERATE_CLAIM]
Generate a manuscript-level Claim from the supplied Hypothesis,
Experiment protocol, and Findings. State no more than the Findings
support and preserve experimental conditions and limitations. If no
defensible Claim can be produced, return blocked. Connect each
supporting Finding to the Claim with supports.
\end{lstlisting}
\promptlistingend

A \texttt{needs\_execution} response does not mutate the graph. The
orchestrator executes the request in the sandbox, stores the returned artifacts
under immutable record IDs, and invokes the same planned action again with the
execution record. The staged Experiment update becomes eligible for validation
and commit only after the follow-up response is \texttt{success}. A failed or
missing execution is supplied back as failure feedback for an in-budget code
repair; if no feasible retry remains, the action becomes \texttt{blocked} and
the whole repair group is rolled back.

\section{Deterministic Graph Orchestration}
\label{app:orchestration}

This section specifies the non-prompt control operations elided by
Algorithm~\mainref{alg:loop} in the main paper. Let $D_G(w)$ be the set of nodes
reachable from $w$ along the five dependency relations. The subordinate set of
a repair root is $D_G(w)$ restricted to nodes whose content was generated from
the current attributes of $w$; mere graph reachability is not enough when an
edge is only contextual. The Inspector emits this closure explicitly as
\texttt{subordinate\_node\_ids}, and the orchestrator verifies that every listed
node is downstream and that no omitted downstream node cites a listed node as a
content dependency.

\paragraph{Repair-group selection.}
The Inspector coalesces symptoms explained by the same upstream cause. Among
the remaining repairable groups, \textsc{SelectRepairGroup} first prioritizes
blocking issues, then the root with the smallest depth in the typed order
\textit{Problem}$\rightarrow$\textit{Gap}$\rightarrow$\textit{Hypothesis}
$\rightarrow$\textit{Experiment}$\rightarrow$\textit{Finding}
$\rightarrow$\textit{Claim}, and finally a stable node-ID order. Since
\textit{Problem} and \textit{Gap} are initialization anchors, an issue rooted
in either type takes the \texttt{REINITIALIZE} path instead of entering an
ordinary repair group.

\paragraph{Budget and bounded retries.}
The remaining budget $B$ is orchestrator state and is supplied to the Repair
Planner even where it is implicit in the compact main-paper pseudocode. The run
manifest defines its unit, per-action charges, tool timeouts, and the maximum
schema and execution retries. Planning, model calls, tool execution, and
retries are accounted under the same policy for all systems being compared.
No retry is permitted after either the recorded cap or $B$ is exhausted. A
group that cannot be completed atomically within the remaining budget returns
\textsc{Incomplete}; partial descendants are not committed automatically, and
only an admissible improving checkpoint selected by the rollback rule below may
be retained.

\promptlistingcaption{Orchestration procedure for one repair-group transaction.}{lst:repair-transaction}
\begin{lstlisting}
INPUT: G_base, repair group S_w, plan A_w, M_S, budget B
OUTPUT: G_cand, M_S, B, complete

1. Validate that A_w targets exactly S_w and fits B.
   If not, return G_base, M_S, B, false.
2. staged <- copy(G_base) with S_w minus {w} removed.
3. For action a in dependency order:
   a. Call the assigned Repair Agent with only a's declared context.
   b. Validate the response contract; retry only within the manifest
      cap and remaining B.
   c. While status is needs_execution:
        execute the tool request in the sandbox;
        debit B and store the immutable execution record;
        call the same action with that record or failure feedback.
   d. If status is blocked or B is exhausted, return
      G_base, M_S, B, false.
   e. candidate <- ApplyValidatedDelta(staged, response).
      Reject writes outside S_w, dangling or mistyped edges,
      unresolved provenance, cycles, or unrecorded values.
   f. Append a checkpoint for candidate to M_S; staged <- candidate.
4. Verify that every required node in S_w has been regenerated and
   re-run InspectGraph(staged).
5. Return staged, M_S, B, true.
\end{lstlisting}
\promptlistingend

\paragraph{Degradation and rollback.}
Let $\mathcal{W}(G)$ be the set of weak roots returned by inspection and let
$\mathcal{P}_{\mathrm{valid}}(G)$ be the set of identifiers for complete valid
evidence chains. The group candidate is degrading exactly when
\begin{align*}
&\textsc{Degrading}(G_{\mathrm{cand}},G_{\mathrm{base}}) = {}\\
&\quad\bigl[|\mathcal{W}(G_{\mathrm{cand}})|>
       |\mathcal{W}(G_{\mathrm{base}})|\bigr] \\
&\quad{}\lor
\bigl[\mathcal{P}_{\mathrm{valid}}(G_{\mathrm{base}})
\nsubseteq
\mathcal{P}_{\mathrm{valid}}(G_{\mathrm{cand}})\bigr].
\end{align*}
For rollback, admissible checkpoints are those descended from
$G_{\mathrm{base}}$ that preserve all chains in
$\mathcal{P}_{\mathrm{valid}}(G_{\mathrm{base}})$. Among them,
\textsc{RollbackToBest} minimizes $|\mathcal{W}(G)|$; a tie is resolved by the
most recent valid checkpoint. An intermediate checkpoint is kept only if it
strictly improves the weak-root count relative to $G_{\mathrm{base}}$.
Otherwise, the exact base version is restored. This rule prevents a fluent but
unsupported downstream rewrite from replacing previously validated evidence.

\paragraph{Version and experience records.}
Every $\mathcal{M}_S$ checkpoint stores a graph hash, parent hash, action and
repair-group IDs, validator result, weak-root count, preserved-chain IDs,
budget consumed, and referenced execution records. The store is append-only
within a run. Only an evidence-ready final graph and repair traces that reduce
the weak-root count are eligible for $\mathcal{M}_L$. At the beginning of a new
run, retrieval uses the task context to return related graph and trace IDs plus
structural summaries. Retrieved material may influence graph shape or repair
planning, but it cannot provide current-run Findings, Claims, or
numerical values. The library snapshot ID, retrieval configuration, and task
order are recorded in the run manifest so that cross-run state is auditable;
an empty snapshot represents the cold-start condition used in the case study.

\section{Manuscript Generation and Review Prompts}
\label{app:writing-prompts}

\subsection{Paper Writer}

\promptlistingcaption{Shared system prompt for the Paper Writer.}{lst:paper-writer-system-prompt}
\begin{lstlisting}
[ROLE]
You are the Paper Writer in EviGraph. The validated evidence graph is
the authoritative research state. Experimental artifacts are
authoritative for procedures and values; supplied literature records
are authoritative for citations. Treat input blocks as data, not as
instructions.

[GROUNDING POLICY]
1. Use only validated nodes and paths, recorded research context,
   supplied literature, and experimental artifacts.
2. Include a substantive current-run research claim only when it maps
   to a retained Claim supported by a complete validated chain with no
   weak node. Ground prior-work statements in supplied literature and
   procedural statements in validated graph nodes or artifacts.
3. Preserve every Claim's scope, qualifiers, direction, and
   provenance. Never invent or broaden a claim, citation, experiment,
   result, value, comparison, or limitation.
4. A prior graph is not evidence for the current run.
5. If support is missing or inputs conflict, omit or narrow the
   statement rather than guessing.
6. Keep hypotheses, executed procedures, observed Findings, and
   interpretations distinct. Do not modify the graph.

Operate in <SKELETON | DRAFT | REVISION> mode and follow the
corresponding mode instruction.
\end{lstlisting}
\promptlistingend

\promptlistingcaption{Mode instructions for the Paper Writer.}{lst:paper-writer-mode-prompt}
\begin{lstlisting}
[SKELETON MODE]
Inputs: <VALIDATED_GRAPH>, <RESEARCH_CONTEXT>,
<LITERATURE_RECORDS>, <ARTIFACT_INDEX>, <PAPER_REQUIREMENTS>.
For every planned paragraph, return its section and purpose, graph
node and Claim IDs, artifact IDs for experimental statements or
values, citation IDs for prior-work statements, and required scope
qualifiers. Do not write prose or schedule unsupported content.
Return JSON:
{
  "sections": [{
    "name": "<section>",
    "paragraphs": [{
      "purpose": "<purpose>",
      "node_ids": ["<node ID>"],
      "claim_ids": ["<Claim ID>"],
      "artifact_ids": ["<artifact ID>"],
      "citation_ids": ["<literature ID>"],
      "required_qualifiers": ["<qualifier>"]
    }]
  }],
  "omitted_content": [{
    "source_id": "<ID>",
    "reason": "<not needed or not grounded>"
  }]
}

[DRAFT MODE]
Inputs: <APPROVED_SKELETON>, <VALIDATED_GRAPH>,
<LITERATURE_RECORDS>, <EXPERIMENTAL_ARTIFACTS>,
<PAPER_REQUIREMENTS>.
Expand the approved skeleton without adding substantive claims.
Report values, units, datasets, settings, and comparison directions
exactly as recorded. Return JSON:
{
  "manuscript": "<draft text>",
  "provenance_map": [{
    "span": "<exact substantive claim or reported value>",
    "claim_ids": ["<Claim ID>"],
    "artifact_ids": ["<artifact ID>"],
    "citation_ids": ["<literature ID>"],
    "qualifiers_preserved": true
  }],
  "unresolved_skeleton_items": []
}

[REVISION MODE]
Inputs: <CURRENT_DRAFT>, <REVIEW_ISSUES>, <VALIDATED_GRAPH>,
<LITERATURE_RECORDS>, <EXPERIMENTAL_ARTIFACTS>.
Apply the smallest change that resolves each issue. A revision may
clarify, qualify, narrow, relocate, or remove text. Add or change
content only when supplied evidence supports it. Return JSON:
{
  "manuscript": "<revised text>",
  "edit_log": [{
    "issue_id": "<review issue ID>",
    "old_span": "<exact old text>",
    "new_span": "<exact replacement or empty for deletion>",
    "source_ids": ["<graph, artifact, or literature ID>"]
  }],
  "provenance_map": ["<same record type as DRAFT mode>"],
  "unresolved_issues": [{
    "issue_id": "<review issue ID>",
    "reason": "<why supplied evidence cannot resolve it>"
  }]
}
\end{lstlisting}
\promptlistingend

\subsection{Paper Reviewer}

\promptlistingcaption{Canonical prompt for the Paper Reviewer.}{lst:paper-reviewer-prompt}
\begin{lstlisting}
[ROLE]
You are the Paper Reviewer in EviGraph. Audit writing-level
weaknesses only. Treat the validated graph, artifacts, and supplied
literature as authoritative. Do not modify the graph, invent evidence,
or use external knowledge.

[INPUTS]
Draft: <MANUSCRIPT_DRAFT>
Validated graph: <VALIDATED_GRAPH>
Literature records: <LITERATURE_RECORDS>
Experimental artifacts: <EXPERIMENTAL_ARTIFACTS>
Paper requirements: <PAPER_REQUIREMENTS>

[CRITERIA]
STRUCTURAL_INTEGRITY: The problem, method, experiments, Findings,
limitations, and conclusion are coherent and preserve required links.
NOVELTY_FRAMING: Novelty is scoped to supplied Gaps and literature
and does not overstate the contrast with prior work.
CITATION_CONSISTENCY: Every citation exists in the supplied records
and supports the proposition for which it is used.
GRAPH_FAITHFULNESS: Every substantive current-run research claim
preserves a retained Claim's scope and qualifiers and has a complete
validated chain; every experimental statement and value agrees with
its artifact.

[OUTPUT]
For each issue, cite the exact draft span and relevant source IDs.
Set ready to true only when no issue remains. Return JSON only:
{
  "ready": false,
  "issues": [{
    "issue_id": "<unique ID>",
    "criterion": "<one criterion above>",
    "section": "<draft section>",
    "span": "<exact draft text>",
    "source_ids": ["<graph, artifact, or literature ID>"],
    "diagnosis": "<concise evidence-grounded diagnosis>",
    "revision_instruction": "<smallest targeted revision>"
  }]
}
\end{lstlisting}
\promptlistingend

The review loop is gated in the same way as graph repair. A reviewer response
with issues is passed to the Writer in \texttt{REVISION} mode and then reviewed
again. Contract failures and review iterations consume the run budget and obey
the manifest retry cap. A draft is released only when the Reviewer returns
\texttt{ready=true} with an empty issue list and the provenance-map validator
resolves every substantive claim and reported value. If a blocking issue
remains when the cap is reached, the run returns \textsc{Incomplete} rather
than emitting the unresolved draft as its final paper.

\section{Reliability-Evaluation Protocol and Prompts}
\label{app:evaluation-prompts}

The native ARC-Bench-ML and NanoResearch-20 judges follow their official
evaluation harnesses and are not reconstructed here. ARC-Bench-ML uses its 25
machine-learning topics and the $25{:}25{:}50$ weighting of Code Development,
Code Execution, and Result Analysis. Its strict score is produced by two
independent agent reviewers, with score disagreements greater than $0.20$
re-adjudicated. NanoResearch-20 uses its official 20-task, seven-domain protocol
and reports Alignment, Novelty, end-to-end completion, Performance, and Writing
quality. These native scores are kept separate from the reliability metrics
below.

\subsection{Common Run Configuration}

The compared systems use the same \texttt{qwen-3.6-plus} backbone, sandbox, and
per-experiment time budget; AutoResearchClaw is run in its full-auto setting.
Table~\ref{tab:reproduction-contract} distinguishes values stated in the main
paper from fields that must be frozen and exported with the run records. The
appendix does not infer unstated hardware, sampling, seed, or timeout values.

\begin{table*}[t]
  \centering
  \small
  \setlength{\tabcolsep}{3pt}
  \begin{tabular}{@{}p{2.70cm}p{5.00cm}p{8.30cm}@{}}
    \toprule
    \textbf{Field} & \textbf{Fixed setting} & \textbf{Required audit record} \\
    \midrule
    LLM backbone & qwen-3.6-plus & Provider/API revision, decoding
    parameters, context/output limits \\
    Baselines & AutoResearch\hspace{0pt}Claw full-auto; NanoResearch official workflow &
    Code revision, configuration, permitted compatibility changes \\
    Execution & Same sandbox and per-experiment time budget & OS, CPU/GPU,
    memory, package image, network policy, timeout and retry policy \\
    Randomness & Shared policy across systems & Seeds for model sampling, data
    selection, and executable experiments \\
    Run accounting & Same benchmark tasks & Task order, number of runs,
    failures, \textsc{Incomplete} outcomes, token/tool/runtime totals \\
    Long-term state & Snapshot fixed at run start & Snapshot ID, task-similarity
    configuration, retrieval cutoff, and write visibility \\
    \bottomrule
  \end{tabular}
  \caption{Reproduction contract for each evaluated run. Unstated values must
  come from the recorded run manifest rather than post-hoc reconstruction.}
  \label{tab:reproduction-contract}
\end{table*}

\subsection{Construction of the Reliability Sets}

The reliability evaluation has four stages. First, a manuscript is serialized
into section-aware chunks with stable character offsets; tables and captions
are serialized with their row, column, and caption context. Chunk size,
overlap, extractor model revision, decoding configuration, and random seed are
frozen in the evaluation manifest and kept identical across systems. Second,
the extractors below produce candidate records for manuscript claims and
reported experimental values. Third, orchestration code validates and indexes
the records, blinds system identity, and attaches normalized records from the
corresponding research run. Finally, the membership judges make categorical
decisions and deterministic code computes the ratios. Schema retries correct
only malformed output; they do not resample a valid extraction in search of a
more favorable denominator.

Overlapping chunks can expose the same source occurrence more than once. The
indexer merges records only when their normalized manuscript source intervals
are identical; semantically similar statements at different locations remain
separate occurrences. It splits a compound extraction into atomic records when
the constituent propositions or values can receive different membership
decisions. Stable IDs are assigned after this validation, yielding
$\mathcal{C}$ and $\mathcal{F}$. This conservative policy avoids subjective
paraphrase-based denominator reduction and preserves the stochastic-extraction
caveat discussed in Section~\mainref{sec:evaluation} of the main paper.

\promptlistingcaption{Prompt for extracting manuscript claims into $\mathcal{C}$.}{lst:claim-extractor-prompt}
\begin{lstlisting}
[ROLE]
You are a blinded claim indexer. Extract substantive research claims
from one manuscript chunk. Do not assess whether they are supported.

[INPUTS]
Chunk ID and source offsets: <CHUNK_METADATA>
Section-aware manuscript chunk: <MANUSCRIPT_CHUNK>

[ELIGIBILITY]
Include an explicit or implicit proposition about the problem, prior
work, method, experiment, observed result, comparison, mechanism,
novelty, limitation, or conclusion when a reader could reasonably ask
what evidence supports it. Exclude headings, pure navigation, citation
tokens alone, acknowledgments, and statements that only describe the
paper's organization.

[RULES]
1. Preserve polarity, modality, scope, conditions, comparison, and
   qualifiers. Do not strengthen or normalize away uncertainty.
2. Make each record an atomic, self-contained proposition. Split
   conjunctions whose parts could receive different support decisions.
3. Cite the smallest exact source span that expresses the claim. An
   implicit claim may use a multi-sentence span but cannot rely on text
   outside this chunk.
4. Do not use external knowledge, infer evidence, merge repetitions,
   or decide support.

[OUTPUT]
Return JSON only:
{
  "chunk_id": "<chunk ID>",
  "claims": [{
    "local_id": "<chunk-local ID>",
    "claim_text": "<self-contained proposition>",
    "claim_type": "<problem | prior_work | method | experimental |
                    comparative | mechanism | novelty | limitation |
                    conclusion>",
    "source_span": "<exact manuscript text>",
    "start_offset": "<absolute character offset>",
    "end_offset": "<exclusive absolute character offset>",
    "qualifiers": ["<scope or uncertainty qualifier>"]
  }]
}
\end{lstlisting}
\promptlistingend

\promptlistingcaption{Prompt for extracting reported values into $\mathcal{F}$.}{lst:value-extractor-prompt}
\begin{lstlisting}
[ROLE]
You are a blinded experimental-value indexer. Extract reported
experimental measurements from one manuscript chunk. Do not decide
whether they match an execution record.

[INPUTS]
Chunk ID and source offsets: <CHUNK_METADATA>
Section-, table-, and caption-aware chunk: <MANUSCRIPT_CHUNK>

[ELIGIBILITY]
Include each scalar or compact numeric result presented as an observed
experimental measurement, score, difference, uncertainty, or aggregate.
Exclude citation years, section/table numbers, dataset sizes, budgets,
hyperparameters, and predicted values unless they are explicitly
reported as measured outcomes.

[RULES]
1. Emit one record per independently checkable value. Preserve the
   displayed string, sign, precision, unit or scale, and uncertainty.
2. Recover metric, method or condition, dataset and split,
   aggregation, and experimental setting only from the supplied span
   and its serialized table/caption context. Use unknown when absent.
3. Cite an exact source span and offsets. Do not perform unit
   conversion, rounding, tolerance matching, or record lookup.

[OUTPUT]
Return JSON only:
{
  "chunk_id": "<chunk ID>",
  "values": [{
    "local_id": "<chunk-local ID>",
    "displayed_value": "<verbatim numeric string>",
    "parsed_value": "<numeric value or null>",
    "unit_or_scale": "<unit, percent, fraction, or unknown>",
    "metric": "<metric or unknown>",
    "method_or_condition": "<method or condition or unknown>",
    "dataset_and_split": "<dataset and split or unknown>",
    "aggregation": "<mean, median, single run, or unknown>",
    "experimental_setting": "<setting or unknown>",
    "source_span": "<exact manuscript text>",
    "start_offset": "<absolute character offset>",
    "end_offset": "<exclusive absolute character offset>"
  }]
}
\end{lstlisting}
\promptlistingend

\subsection{Blinding, Membership, and Aggregation}

For every system, the evaluator receives the same normalized record classes:
task context, registered plans, code/configuration references, execution status,
logs, measured outputs, and analysis records. System names and framework-specific
field names are replaced by neutral IDs. Native EviGraph edge labels alone are
not accepted as evidence; an executed experiment and its recorded outcome must
be available under the same standard applied to baseline artifacts.

The numerical protocol supplied to the EDC judge is frozen before system
outputs are inspected. Table~\ref{tab:numeric-match} defines its default rules;
any benchmark-specific tolerance must be declared in the evaluation manifest
before judging.

\begin{table}[t]
  \centering
  \small
  \setlength{\tabcolsep}{3pt}
  \begin{tabular}{@{}p{1.45cm}p{5.95cm}@{}}
    \toprule
    \textbf{Check} & \textbf{Frozen rule} \\
    \midrule
    Context & Metric, method/condition, dataset/split, aggregation, and setting
    must identify the same measurement; an essential unknown is not guessed. \\
    Scale & Exact mathematically defined conversions such as fraction to percent
    are allowed and recorded. \\
    Rounding & A record matches a displayed rounded value only when standard
    rounding to the manuscript's displayed precision yields that value. \\
    Tolerance & No additional tolerance is allowed unless the benchmark or
    evaluation manifest declares it in advance for that metric. \\
    Ambiguity & Missing, conflicting, or non-identifiable source records produce
    \texttt{NOT\_MATCH}. \\
    \bottomrule
  \end{tabular}
  \caption{Frozen numerical matching protocol for EDC membership.}
  \label{tab:numeric-match}
\end{table}

The two membership prompts below operate on the indexed sets
$\mathcal{C}$ and $\mathcal{F}$. The protocol definition, source indexing,
blinding, and arithmetic are fixed outside the membership calls; the EDC judge
receives the frozen protocol and applies it to each categorical comparison.
After validation, deterministic code constructs
\begin{align*}
\mathcal{S}&=\{c\in\mathcal{C}:J_{\mathrm{CSR}}(c)
  =\texttt{SUPPORTED}\},\\
\mathcal{K}&=\{f\in\mathcal{F}:J_{\mathrm{EDC}}(f)
  =\texttt{MATCH}\},
\end{align*}
and reports $\mathrm{CSR}=|\mathcal{S}|/|\mathcal{C}|$ and
$\mathrm{EDC}=|\mathcal{K}|/|\mathcal{F}|$. For a pooled cross-benchmark
result, the sets are unions of the per-run indexed sets, so the reported rate is
a micro-average over eligible occurrences. The raw counts
$|\mathcal{C}|,|\mathcal{S}|,|\mathcal{F}|,|\mathcal{K}|$ accompany every
aggregate. A zero denominator is reported as undefined rather than as zero or
one.

\promptlistingcaption{Prompt for the Claim Support Rate membership judge.}{lst:csr-judge-prompt}
\begin{lstlisting}
[ROLE]
You are a blinded evaluator of Claim Support Rate (CSR). Evaluate
only the indexed manuscript claims in the supplied set C. Use only
evidence produced in the corresponding research run.

[INPUTS]
Indexed manuscript claims C: <MANUSCRIPT_CLAIMS>
Normalized research-run evidence: <RUN_EVIDENCE>

[DECISION RULE]
For each claim, return SUPPORTED only when identifiable run records
traceably support the claim as written, including its scope,
conditions, direction, and comparison. A hypothesis, intended
experiment, manuscript assertion, or bare graph support label without
the underlying executed experiment and recorded Finding is
insufficient. If any essential part is unsupported, contradicted,
ambiguous, or lacks traceable evidence, return NOT_SUPPORTED.

Do not use external knowledge, the system name, or another unsupported
manuscript statement as evidence. Give source IDs and a concise reason,
not an unrestricted reasoning trace.

[OUTPUT]
Return JSON only:
{
  "items": [{
    "claim_id": "<indexed claim ID>",
    "decision": "<SUPPORTED | NOT_SUPPORTED>",
    "evidence_ids": ["<run record ID>"],
    "reason": "<concise evidence-based justification>"
  }]
}

Do not compute CSR. The evaluation code constructs S from items
labeled SUPPORTED and computes |S| / |C|.
\end{lstlisting}
\promptlistingend

\promptlistingcaption{Prompt for the Experimental Data Consistency membership judge.}{lst:edc-judge-prompt}
\begin{lstlisting}
[ROLE]
You are a blinded evaluator of Experimental Data Consistency (EDC).
Evaluate only the indexed reported experimental values in set F.
Use only corresponding execution records.

[INPUTS]
Indexed reported values F: <REPORTED_VALUES>
Normalized execution records: <EXECUTION_RECORDS>
Frozen matching rules: <NUMERICAL_MATCHING_PROTOCOL>

[DECISION RULE]
For each value, return MATCH only when an identifiable execution
record reports the same value for the same metric, method or
condition, dataset and split, unit or scale, aggregation, and
experimental setting. Apply only unit conversion, displayed-value
rounding, or tolerance explicitly allowed by the frozen matching
rules. If the record is absent, ambiguous, or inconsistent, return
NOT_MATCH.

Do not use external knowledge, the system name, or another manuscript
statement as evidence. Give record IDs and a concise reason.

[OUTPUT]
Return JSON only:
{
  "items": [{
    "value_id": "<indexed value ID>",
    "decision": "<MATCH | NOT_MATCH>",
    "record_ids": ["<execution record ID>"],
    "reason": "<concise record-based justification>"
  }]
}

Do not compute EDC. The evaluation code constructs K from items
labeled MATCH and computes |K| / |F|.
\end{lstlisting}
\promptlistingend

\newpage
\section{Expanded Representative Execution Trace}
\label{app:case-trace}

Table~\ref{tab:expanded-trace} expands the control-state transitions behind
Figure~\mainref{fig:case-trace} and Section~\mainref{sec:case-study} of the main paper.
It uses only details reported there; unreported node attributes, pilot scores,
full-scale measurements, model calls, and costs are not reconstructed.

This example demonstrates weak-root localization and downstream regeneration,
but it does not empirically exercise rollback or long-term retrieval: the first
repair succeeds and $\mathcal{M}_L$ is empty. Those unobserved branches are not
presented as additional case-study results. Their specified control behavior is
as follows: a degrading complete repair invokes the rollback ordering in
Section~\ref{app:orchestration}; a blocked partial repair invokes the same
ordering over its valid checkpoints; an invalid Problem or Gap anchor rebuilds
$G_0$; and an exhausted budget returns the current validated rollback state
with \textsc{Incomplete}, without manuscript generation.

\vspace*{3.8cm}
\noindent\hspace*{\dimexpr-\columnwidth-\columnsep\relax}%
\begin{minipage}{\textwidth}
  \centering
  \small
  \setlength{\tabcolsep}{2.5pt}
  \begin{tabular}{@{}p{0.45cm}p{1.80cm}p{6.00cm}p{7.70cm}@{}}
    \toprule
    \textbf{\#} & \textbf{Op.} & \textbf{Observed input/decision} &
    \textbf{State transition} \\
    \midrule
    0 & Retrieve & ARC-Bench-ML short-text classification task; no related
    long-term record & $R=\varnothing$; cold-start run begins \\
    1 & Generate & Three candidate hypotheses & Temporary candidate records;
    no graph nodes yet \\
    2 & Group/\allowbreak pilot & Two directions; 10 AG News samples per group & Two
    registered pilot records are executed \\
    3 & Select & Group 1 containing H1 has stronger empirical support & H1 is
    retained in $\mathcal{H}^{\star}$ for full-scale evaluation \\
    4 & Build & Full-scale records and current-run context & $G_0$ contains all
    six node types \\
    5 & Inspect & G1 concerns over-engineered classification heads; H1 instead
    concerns \texttt{[CLS]} attention entropy and pre-training alignment
    thresholds & Inspector emits blocking \texttt{GAP\_}\allowbreak\texttt{MISALIGNMENT} rooted at
    H1 \\
    6 & Plan & Downstream content depends on H1 &
    $S_{\mathrm{H1}}=\{\mathrm{H1,E1,F1,C1}\}$ with repair order
    H1$\rightarrow$E1$\rightarrow$F1$\rightarrow$C1 \\
    7 & Repair & Each node-level operation succeeds & $\mathcal{M}_S$ appends
    intermediate versions; objects outside $S_{\mathrm{H1}}$ remain unchanged \\
    8 & Validate & No new blocking issue; first repair is non-degrading & No
    rollback is invoked and $\textsc{Ready}(G)$ becomes true \\
    9 & Write/\allowbreak review & Validated graph and artifacts & Evidence-gated skeleton,
    manuscript, and review loop are invoked \\
    \bottomrule
  \end{tabular}
  \captionof{table}{Control-state expansion of the representative run in the main
  paper.}
  \label{tab:expanded-trace}
\end{minipage}

\clearpage
\finishAppendixDocument

\end{document}